\documentclass[letterpaper,twocolumn,10pt]{article}
\usepackage{usenix}

\usepackage{amssymb,amsfonts}
\usepackage{graphicx}
\usepackage{textcomp}
\usepackage{color}
\usepackage{algorithm}
\usepackage{caption}
\usepackage{algpseudocode}
\usepackage{amsmath}
\floatname{algorithm}{Algorithm}

\usepackage{xspace}

\DeclareMathOperator*{\Max}{max}

\newcommand{\sysname}{\emph{Refiner}\xspace}

\newcommand{\revise}[1]{\textcolor{black}{#1}}

\usepackage[normalem]{ulem}
\usepackage{enumitem}
\usepackage{array}
\usepackage{diagbox}
\newcommand{\tabincell}[2]{\begin{tabular}{@{}#1@{}}#2\end{tabular}}

\usepackage{epstopdf}
\usepackage{hyperref}
\usepackage{booktabs}
\usepackage{multirow}
\usepackage{soul}
\usepackage{float}
\usepackage{subfigure}
\usepackage[numbers,sort&compress]{natbib}

\newcommand{\ie}{\textit{i.e.}}

\newtheorem{thm}{\bf Theorem} 
\newtheorem{assumption}{Assumption} 
\newtheorem{observation}{\bf Insight} 
\newtheorem{example}{\bf Example} 
\usepackage{tikz}
\usepackage{amsmath}

\usepackage{filecontents}
\usepackage{xcolor}

\begin{document}

\date{}

\title{\sysname: Data Refining against Gradient Leakage Attacks in Federated Learning}

\author{{\rm Mingyuan Fan$^1$} \hspace{5mm} {\rm Cen Chen$^{1,2}$}\thanks{Corresponding author.} \hspace{5mm} {\rm Chengyu Wang$^{3}$} \hspace{5mm} {\rm Xiaodan Li$^{1,3}$} \hspace{5mm} {\rm Wenmeng Zhou$^{3}$} \\
$^1$East China Normal University \\
$^2$The State Key Laboratory of Blockchain and Data Security, Zhejiang University \\
$^3$Alibaba Group \\
mingyuan\_fmy@stu.ecnu.edu.cn \hspace{5mm} cenchen@dase.ecnu.edu.cn \hspace{5mm} \\
\{ chengyu.wcy, fiona.lxd, wenmeng.zwm\}@alibaba-inc.com
}

\maketitle

\begin{abstract}
    Recent works highlight the vulnerability of Federated Learning (FL) systems to \textit{gradient leakage attacks}, where attackers reconstruct clients' data from shared gradients, undermining FL's privacy guarantees.
    However, existing defenses show limited resilience against sophisticated attacks.
    This paper introduces a novel defensive paradigm that departs from conventional gradient perturbation approaches and instead focuses on the construction of robust data.
    Our theoretical analysis indicates such data, which exhibits low semantic similarity to the clients' raw data while maintaining good gradient alignment to clients' raw data, is able to effectively obfuscate attackers and yet maintain model performance.
    We refer to such data as robust data, and to generate it, we design \sysname that jointly optimizes two metrics for privacy protection and performance maintenance.
    The utility metric promotes the gradient consistency of key parameters between robust data and clients' data, while the privacy metric guides the generation of robust data towards enlarging the semantic gap with clients' data.
    Extensive empirical evaluations on multiple benchmark datasets demonstrate the superior performance of \sysname at defending against state-of-the-art attacks.
\end{abstract}

\section{Introduction}
\label{intro}
\begin{figure}[!t]
    \centering
    \includegraphics[width=0.45\textwidth]{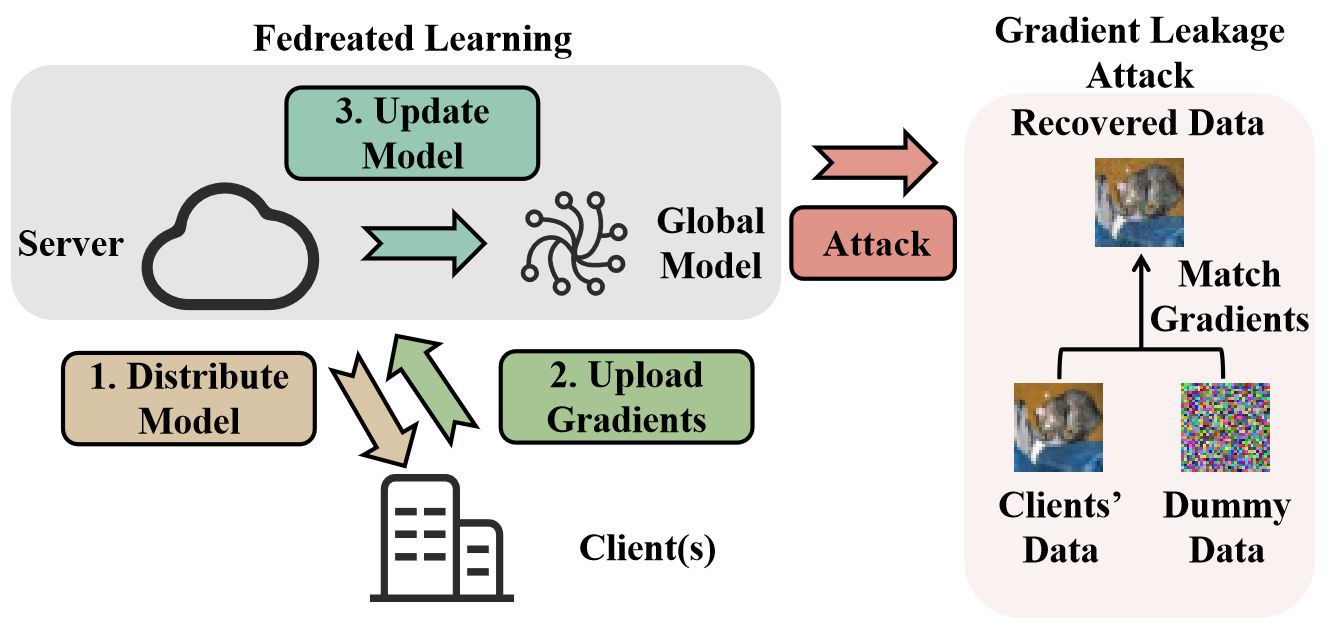}
    \caption{A sketch map of FL and GLAs.}
    \label{fed_gla_scenario}
\end{figure}

\revise{Federated Learning (FL) enables privacy-preserving DNN training by having clients compute gradients locally and share them with a central server for updating the global model~\cite{fan2025trustworthiness}, as depicted in Figure \ref{fed_gla_scenario}.
However, recent \textit{Gradient Leakage Attacks} (GLAs)~\cite{gla,grad_inversion} have shown that clients' data can be reconstructed by aligning uploaded gradients with those generated from dummy data.
While some works \cite{du2024sok,wang2024more} argued that such attacks are limited to simple scenarios, recent advances in GLAs \cite{grad_inversion,gan_prior} have shown these threats are increasingly effective in practical FL settings, especially when the server is allowed to manipulate the global model \cite{ZhaoSEEAB24,model_inconsistency}.
Moreoverr, the uploaded gradients can also be utilized to infer dataset properties and membership information \cite{RecUP,FLMIA}.
This highlights the need for effective defenses in FL to safeguard the confidentiality of sensitive client information.}

Two primary types of methods have been proposed to safeguard privacy in FL:
encryption-based methods~\cite{CryptoNets,BatchCrypt,Safelearn,secure_agg} and perturbation-based methods~\cite{dp,gla,grad_inversion_survey,soteria,soteria2}.
In encryption-based methods~\cite{Safelearn}, clients transmit encrypted gradients, but these encrypted gradients are ultimately decrypted into plaintext for aggregation purposes~\cite{Safelearn,secure_agg}, leaving them vulnerable to GLAs~\cite{gla}.
Moreover, encryption-based methods incur significant computational/communication overheads that dominate the training process, significantly affecting their practicability~\cite{soteria2,BatchCrypt}.

Perturbation-based methods impose subtle perturbations into ground-truth gradients, which helps to obscure the adversary’s (the server’s) ability to recover accurate data from the gradients.
Notable defenses include differential privacy~\cite{dp} and gradient pruning~\cite{grad_pruning}, which perturb gradients by injecting random noise or discarding part of gradients. 
State-of-the-art defenses~\cite{soteria,soteria2} estimate the privacy information contained in each gradient element for gradient pruning.
\revise{Despite demonstrating promising privacy protection ability in certain FL scenarios, these methods inevitably compromise the model's utility due to gradient perturbations.
Consequently, developing methods that can achieve a better trade-off between utility and privacy remains a significant research topic.}

\begin{figure}[t!]
    \centering
    \includegraphics[width=0.45\textwidth]{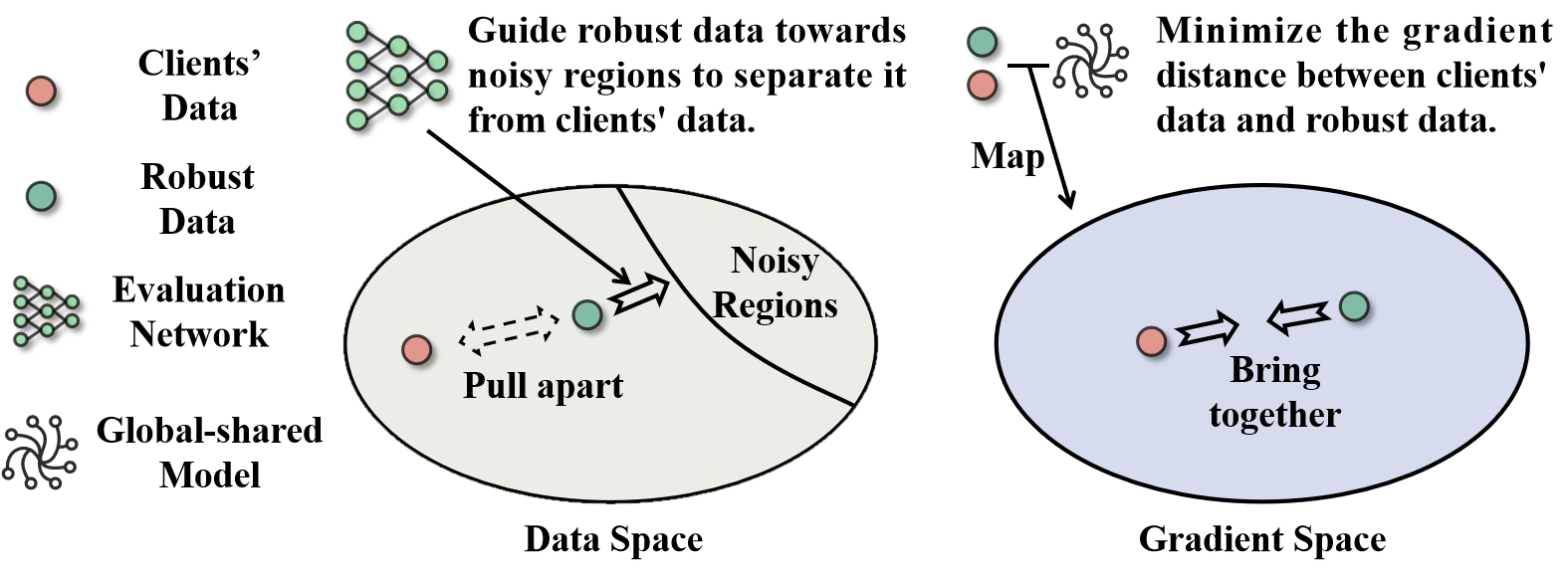}
    \caption{
    \sysname enhances privacy protection by employing an evaluation network that directs robust data toward noisy regions, increasing the semantic distance from clients' data. To maintain utility, \sysname utilizes the global-shared model to map clients' data and robust data onto gradient space and minimizes the gradient distance between them.
    }
    \label{overview}
\end{figure}
\textbf{Our contributions.}
This paper explores a novel defensive paradigm that diverges from conventional gradient perturbation approaches, centering instead on the construction of robust data.
Our inspiration stems from an intuitive yet compelling idea: if we can generate robust data that exhibit substantial semantic dissimilarity compared to clients' data while maintaining utility, the gradients associated with these data are likely to have the potential to effectively obfuscate attackers while imposing only a small degradation in the model's performance.
To bring this idea to fruition, we present \sysname which crafts robust data by jointly optimizing two key metrics for privacy protection and performance maintenance.
As illustrated in Figure \ref{overview}, the privacy metric, \ie, the evaluation network, encourages robust data to deviate semantically from clients' data, while the utility metric aligns the gradients of model parameters associated with clients' data and robust data.
In practice, to protect privacy, clients refrain from uploading gradients directly derived from their raw data.
Instead, they upload gradients associated with robust data, thereby maintaining the confidentiality of clients' data.
\begin{itemize}[leftmargin=*,topsep=0pt,itemsep=-5pt]
\item \textit{Utility metric.}
    The utility metric is defined as the weighted distance between the gradients of model parameters with respect to robust data and clients' data.
    By minimizing the gradient distance, the utility of the robust data approaches that of the clients' data.
    The weighting scheme prioritizes the preservation of gradients associated with important parameters, thus better maintaining the utility of the robust data. 
    This weighting mechanism consists of element-wise weight and layer-wise weight.
    The element-wise weight is determined based on empirical and theoretical observations, considering the values of both gradients and parameters to identify critical parameters. 
    Additionally, the layer-wise weight exploits the significance of earlier layers by allocating more attention to them during the learning process.
\item \textit{Privacy metric.}
    While norm-based distances between robust data and original data are commonly used to measure privacy~\cite{soteria}, their effectiveness is limited due to the significant divergence with human-vision-system difference metrics~\cite{metric}. 
    To address this limitation, \sysname instead employs an evaluation network to accurately estimate the human perception distance between robust data and original data. The \textit{perception distance} serves as the privacy metric, reflecting the overall amount of disclosed privacy information associated with clients' data. 
    By leveraging the human perception distance, \sysname provides a more accurate and effective evaluation of the privacy information contained in robust data.
\end{itemize}
\revise{The main body of this paper focuses on the threat of data reconstruction due to its severity, while discussions regarding \sysname's performance on text data and other types of privacy leakage are included in Appendix \ref{appendix_effect_mia_aa}.}

\section{Background \& Related Work}
\label{related_work}

\subsection{Federated Learning}

This paper is anchored in a typical FL scenario, where a server collaborates with $m$ clients to train a model $F_{\theta}(\cdot)$ parameterized by $\theta$, utilizing loss function $\mathcal{L(\cdot, \cdot)}$.
The $i$-th client's local dataset is denoted by $D_i$.
The server and clients interact with each other over $N$ communication rounds by executing the two steps:
\begin{itemize}[leftmargin=*,topsep=0pt,itemsep=-3pt]
\item \textbf{Model Distribution and Local Training.}
The server samples a subset of $K$ clients from the entire client pool and broadcasts the global model $F_{\theta}(\cdot)$ to selected clients.
The $k$-th selected client ($k=1,\cdots,K$) computes gradients $g_k=\nabla_{\theta} \mathcal{L}(F_{\theta}(x),y)$ using raw data $x,y$ sampled from its local dataset.
\item \textbf{Global Aggregation.}
The server averages the gradients received from the selected clients to update the global model: $\theta = \theta -  \frac{\eta}{K} \sum_{i=1}^{K} g_i$ where $\eta$ is the learning rate.
\end{itemize}

In the literature~\cite{fed_quant}, clients may update the local model over multiple steps and upload the model parameters.
We employ a default local step of 1, where clients directly upload the gradients.
This decision is motivated by the fact that the server typically derives gradients by comparing the global model parameters with the uploaded local model parameters.
Multiple local steps could potentially cause a discrepancy between the derived gradients and the ground-truth gradients, thereby reducing attack effectiveness~\cite{grad_inversion_survey}.
Thus, we adopt the setting with the most powerful attack scenario to better examine defense performance.

\subsection{Threat Model}

We consider a prevalent attack scenario in which the adversary is an \textit{honest-but-curious} server~\cite{soteria2}.

\begin{itemize}[leftmargin=*,topsep=0pt,itemsep=-3pt]
\item \textbf{Adversary's Objective.} The primary goal of the server is to recover private information from the gradients shared by clients, endeavoring to capture a wealth of semantic details involved in clients' raw data.
\item \textbf{Adversary's Capabilities.} The server will faithfully execute FL protocols. The server can access the model's weights and the uploaded gradients but is unable to manipulate the model's architecture or weights to better suit their attacks.
\item \textbf{Adversary's Knowledge.} The server can leverage public datasets to augment their attacks and possesses sufficient computational resources. These public datasets could be similar to clients' data to serve as strong priors for enhancing their attacks, but are not duplicates of the clients' datasets.
\end{itemize}

\revise{In addition, we also study a malicious active server, where the adversary can manipulate model parameters during training.
Appendix \ref{eval_loki} evaluates \sysname's performance under the stronger malicious threat model.
Notice that, the honest-but-curious server remains the primary focus of this paper due to its practical prevalence in real-world FL deployments.}
We suppose that clients are aware of the privacy risks, and, as a remedy, clients proactively adopt defenses.

\begin{itemize}[leftmargin=*,topsep=0pt,itemsep=-3pt]
    \item \textbf{Defender's Objective.} Clients' objective is to prevent the server from reconstructing their data by using the gradients they upload. Moreover, clients wish to maintain model performance.
    \item \textbf{Defender's Capabilities.} Clients possess the autonomy to adjust their ground-truth gradients before upload, utilizing moderate computational resources to enable their defense strategies.
    \item \textbf{Defender's Knowledge.}
    Defender's knowledge includes their awareness of the server's reliability and possible countermeasures.
    We assume that clients know potential defense methods and can access to necessary resources from trusted third parties for implementing these defenses, such as evaluation network (See Section \ref{privacy_metric}). 
\end{itemize}

To simplify the notations, subscripts, and superscripts representing specific clients are omitted throughout the paper.

\subsection{Gradient Leakage Attack}

Given randomly initialized dummy data $x'$ with dummy label $y'$ and uploaded gradients $g$, the seminal GLA \citet{gla} frames reconstruction as a gradient matching problem:
\begin{equation}
\label{reconstruction_process}
\begin{split}
    x',y' = {\arg \min}_{x',y'} ||\nabla_{\theta} \mathcal{L}(F_{\theta}(x'),y')-g||_2.
\end{split}
\end{equation}
The follow-up works~\cite{label_leakage,igla} showed how to infer the ground-truth label $y$ from the gradients of the fully-connected layer, reducing the complexity of Equation \ref{reconstruction_process}.
However, several studies~\cite{evaluate_gla,invert_grad,grad_inversion} empirically demonstrated that bad initialization causes convergence towards local optima, making recovered data noisy.
In response, \citet{invert_grad} and \citet{gan_prior} explored various regularization techniques and heuristic tricks, including total variation, restart, and improved initialization, aiming to eliminate invalid reconstructions.
Recently, since the output of generative models is not noisy, \citet{gga} and \citet{grad_obfuscation} trained a generative network to recover clients' data.
The potential assumption is the availability of an abundance of data that is similar to the clients' data.
They optimized latent vectors of the generative network to produce images whose gradients are similar to uploaded ones.

\subsection{Gradient Leakage Defense}

While cryptographic methods have emerged as potential solutions in defending GLAs, the substantial overhead restricts their applicability~\cite{gla,soteria2,gga} and motivates more lightweight perturbation-based defenses.
One such perturbation-based method~\cite{gla,bayes_attack_for_gla,dp_gla} is differential privacy (DP), which adds random noises to ground-truth gradients to generate perturbed gradients.
Moreover, some works~\cite{gla,grad_obfuscation} explored gradient pruning~\cite{grad_pruning} or gradient quantization (GQ)~\cite{fed_quant} to obfuscate the server.
Soteria~\cite{soteria} employs a theoretically-derived metric to evaluate the privacy information contained in each gradient element and selectively prunes gradients in fully-connected layers.
However, \citet{bayes_attack_for_gla,soteria2} identified that Soteria could be compromised by muting gradients in fully-connected layers during the reconstruction process (Equation~\ref{reconstruction_process}).
In response, they introduced a gradient mixing mechanism to solve the problem.

\subsection{Image Anonymization}

Robust data is defined to be differently semantic from clients' data while maintaining comparable utility.
Image anonymization strips sensitive content from images, thereby offering a potential approach for generating robust data.
Traditional anonymization techniques include blurring, masking~\cite{newton2005preserving}, and k-means clustering~\cite{image_anony_kmeans,image_anony_kmeans2}.
Recent advancements~\cite{modern_image_anony,modern_image_anony2} introduce the use of generative models for the synthetic replacement of sensitive content.
However, image anonymization has been demonstrated to be vulnerable to malicious attacks~\cite{neustaedter2006blur,gallagher2008clothing,zhang2015beyond,oh2016faceless}, causing potential privacy leakage.
Moreover, these methods are primarily developed with privacy protection, instead of finding a satisfactory trade-off between utility and privacy~\cite{hukkelaas2023does}.
As a result, directly employing image anonymization in \sysname is unsuitable.

\section{Our Approach: \sysname}
\label{approach}

\subsection{Problem Formulation}

\sysname aims to construct robust data $x^*$ that preserves high utility while minimizing privacy information leakage from raw client data $x$.
Then, the gradients generated by $x^*$ serve as uploaded gradients.
\sysname formulates the following optimization problem to craft robust data:
\begin{equation}
\label{optim_task}
    x^*=\mathop{\arg \min}_{x^*} UM(x^*,x) - \beta \cdot PM(x^*,x),
\end{equation}
where $UM$ measures the utility gap between $x^*$ and $x$ (lower values indicate better utility), and $PM$ quantifies the reduction in privacy information leakage relative to $x$ (higher values indicate stronger privacy).
$\beta$ is a balance factor to regulate the trade-off between utility and privacy.
A higher value of $\beta$ intensifies the prioritization of privacy.
Section \ref{utility_metric} and Section \ref{privacy_metric} will elaborate on $UM$ and $PM$, respectively.

\subsection{Utility Metric}
\label{utility_metric}

\subsubsection{Motivation}

The utility of data is defined as the extent to which it contributes to reducing the model loss.
DNNs utilize gradients to update model parameters to minimize loss, making it straightforward to exploit the mean square error (MSE) between gradients generated by $x$ and $x^*$ as the utility metric, \ie, $UM(x^*,x)=|| \nabla_{\theta} \mathcal{L}(F_{\theta}(x^*),y) - \nabla_{\theta} \mathcal{L}(F_{\theta}(x),y)  ||_2^2$.
However, employing the MSE in its vanilla form treats every gradient element without distinction, neglecting that some parameters may exert a more pronounced influence.

To address this, we customize a weighted MSE to more effectively evaluate the utility contained in $x^*$.
The overall idea of weighted MSE is to endow the gradients of important parameters with higher weights, so as to pay more attention to aligning the gradients of important parameters associated with $x$ and $x^*$.
While existing parameter importance estimation techniques exist, they often incur high computational overhead (See Appendix \ref{weighting_mechanism_comparison}).
We instead design a lightweight weighting mechanism that trades off a slight performance compromise for considerable savings in computational costs.

To determine the weights, we consider two factors: element-wise weight along with layer-wise weight.
The product of element-wise weight and layer-wise weight is used as the ultimate weight for the corresponding gradient element.

\subsubsection{Element-wise Weight}

\begin{observation}
\label{obs1}
\textit{
    The significance of parameters is closely tied to both their values and gradients.
    Intuitively, parameters with high-magnitude values are crucial since they can considerably amplify their inputs, thereby yielding a high output that is more likely to exert substantial influence in the model’s predictions. 
    Besides, the essence of gradient is to quantify the sensitivity of the loss function to small changes in the parameter value~\cite{convex_opt}.
    Hence, parameters associated with larger gradients are considered more influential as they have a greater impact on shaping the model performance.
}
\end{observation}

Insight \ref{obs1} motivates defining element-wise weights as the absolute value resulting from multiplying the parameter's value with corresponding gradients.
The values of parameters reflect their significance to upstream layers, and the gradients of parameters suggest their importance to downstream layers.
Consequently, the element-wise weight intuitively serves as an effective metric to evaluate the importance of parameters.

\textbf{Theoretical justification.}
We theoretically validate the effectiveness of element-wise weight using the $Q$-function $Q(u) = \mathcal{L}(F_{u\theta}(x), y)$, where $u$ scales parameter magnitudes~\cite{bayes_book}.
Taking the derivative of $Q(u)$ with respect to $u$ and setting $u=1$ yields $Q'(1) = \nabla_{\theta} \mathcal{L}^T \theta$.
It is well-known that the model achieves its optimality when $\nabla_{\theta} \mathcal{L}=0$~\cite{convex_opt}.
Therefore, the value of $Q'(1)$ can be used to check the model's optimality.
Enforcing $Q'(1) \approx 0$ requires minimizing $|Q'(1)| = |\sum_i \nabla_{\theta_i} \mathcal{L} \cdot \theta_i|$, where $i$ indicates the $i$-th element of the vector.
If only a subset of parameters is allowed to be optimized, focusing on optimizing the parameters with the largest product of absolute weight and gradient, \ie, the parameters that our element-wise weight identifies, can minimize $Q(1)$ to the fullest extent.

\begin{table}[]
\centering
\caption{The model performance with three different gradient pruning strategies.}
\label{element_metric_valid}
\scriptsize
\begin{tabular}{@{}ccccc@{}}
\toprule
Pruning Rate & 0.2 & 0.4 & 0.6 & 0.8 \\ \midrule
Weight & 51.35 & 48.61 & 46.80 & 42.72 \\
Grad & 51.67 & 50.90 & 49.14 & 47.35 \\
Ours & \textbf{52.88} & \textbf{51.45} & \textbf{50.96} & \textbf{49.54} \\ \bottomrule
\end{tabular}
\end{table}

\textbf{Empirical validation.}
We examine element-wise weight against gradient- and weight-based pruning on LeNet trained on CIFAR-10 (20 epochs, learning rate 0.01, batch size 128), following \citet{gla}.
We investigate the performance implications of three distinct gradient pruning strategies: (1) pruning based on the smallest absolute gradient (Grad), (2) the smallest absolute weight (Weight), and (3) the smallest absolute product of weight and gradient (Ours).
Table \ref{element_metric_valid} compares test accuracy under varying pruning ratios.
As can be seen, element-wise weight achieves the best performance, as it can better identify important parameters compared to the other two pruning strategies.

\subsubsection{Layer-wise Weight}

\begin{observation}
\label{obs2}
\textit{
    DNNs exhibit hierarchical parameter importance, with shallow layers playing a more critical role in model performance. This stems from two key properties: (1) perturbations in shallow layers propagate exponentially through forward propagation, amplifying their impact on outputs; and (2) shallow layers specialize in learning low-level, shared features across data samples~\cite{Simonyan2014VeryDC}.
}
\end{observation}

\begin{table}[]
\centering
\caption{The model performance when perturbing different layers. The settings are the same as those in Table \ref{element_metric_valid}.}
\label{layer_metic_valid}
\scriptsize
\begin{tabular}{@{}ccccc@{}}
\toprule
Variance & Layer 1 & Layer 2 & Layer 3 & Linear Layer \\ \midrule
0.001 & 48.94 & 49.92 & 50.84 & 51.73 \\
0.01 & 48.40 & 49.72 & 50.25 & 51.60 \\
0.1 & 47.56 & 49.55 & 49.54 & 51.38 \\ \bottomrule
\end{tabular}
\end{table}

\begin{figure}[!t]
	\centering
	\subfigure[Original]{
	\includegraphics[width=0.15\linewidth]{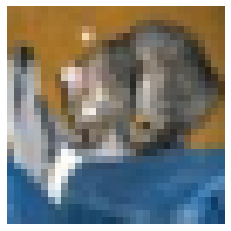} }
	\subfigure[Shift]{ \includegraphics[width=0.15\linewidth]{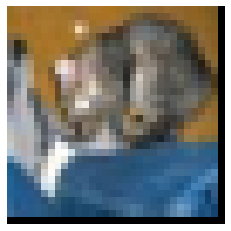} }
	\subfigure[Scale]{ \includegraphics[width=0.15\linewidth]{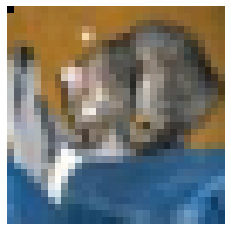} }
	\subfigure[Noise]{ \includegraphics[width=0.15\linewidth]{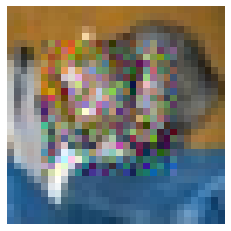} }
	\caption{
        We perturb the original image by shifting it towards the top-left direction by a unit, scaling a pixel, and adding random noises.
        These make MSEs of 0.025, 0.106, and, 0.004, respectively.
        Adding random noises produces better privacy protection visually than shifting and scaling, but the former two yield higher MSEs.
        See Appendix \ref{validation_ssim_lpips} for SSIM and LPIPS.
         }
	\label{example_for_metric_shortcoming}
\end{figure}

Drawing upon Insight \ref{obs2}, the parameters in shallow layers commonly are more important than those in late layers.
As reported in Table \ref{layer_metic_valid}, empirical validation using Gaussian noise injection in gradient confirms that disrupting shallow-layer gradients causes significantly greater performance degradation than deeper layers, motivating our layer-wise weight.

The specific form of layer-wise weight draws inspiration from the multiplicative structure of DNNs.
DNNs typically consist of stacked layers, each of which is a linear function followed by a nonlinear activation function $\sigma(\cdot)$~\cite{deep_learning_survey}.
In mathematical terms, we represent a DNN as $ \sigma \cdots \sigma (w_2 \sigma (w_1 x)) $.
This structure implies that a perturbation in the parameters of a single layer can exert an exponential effect on the final model output, specifically when considering piece-wise linear activation functions like ReLU.
Hence, our layer-wise weight employs an exponential decay mechanism to assign weights.

Suppose that $F_{\theta}(\cdot)$ comprise a total of $K$ layers, with the $i$-th layer parameterized with $\theta[i]$ ($\theta = \{\theta[1], \theta[2], \cdots, \theta[K]\}$).
The layer-wise weight of gradient elements in $i$-th layer $\theta[i]$ is defined as $power(\tau, i)$, where $\tau$ is a decay factor ($0 \leq \tau \leq 1$) and $power(\cdot, \cdot)$ is the power function.

\subsubsection{Ultimate Weight}

We define the ultimate weight for $\theta[i]$ as the product between element-wise weight and layer-wise weight:
\begin{equation}
    weight(\theta[i]) = |grad(\theta[i]) \cdot value(\theta[i]) \cdot power(\tau,i)|,
\end{equation}
where $grad(\theta[i])$ and $value(\theta[i])$ denote taking out the gradients associated with $\theta[i]$ and values of $\theta[i]$.
$UM$ is then represented as the weighted sum of the squared differences between the gradients associated with $x$ and $x^*$:
\begin{equation}
\begin{split}
        UM(x^*,x)&=||\sum_{i=1}^{K} weight(\theta[i]) \\ &(\nabla_{\theta[i]} \mathcal{L}(F_{\theta}(x^*), y) - grad(\theta[i]))||_2^2.
\end{split}
\end{equation}

\subsection{Privacy Metric}
\label{privacy_metric}
\subsubsection{Motivation}

\begin{figure}[!t]
    \centering
    \includegraphics[width=0.6\linewidth]{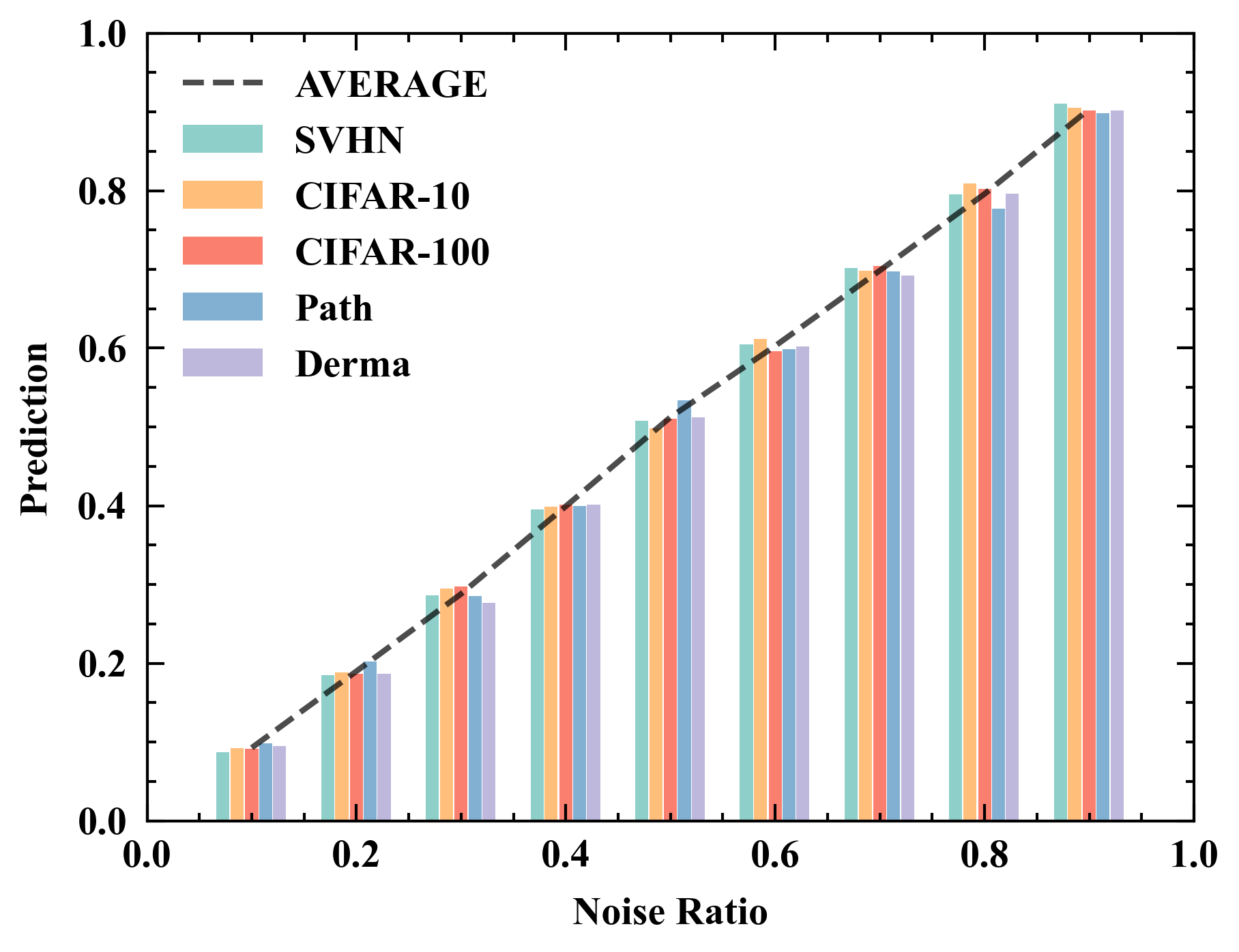}
    \caption{We extract 1000 samples from SVHN, CIFAR10, CIFAR100, Path, and Derma. We report the average predictions of the evaluation network for the mixture of these samples and random noises with proportions ranging from 0.1 to 0.9.}
    \label{evaluation_network_nums1}
\end{figure}

PM aims to quantify the level of privacy leakage regarding $x$ caused by $x^*$, or in other words, it measures the dissimilarity between $x$ and $x^*$ in terms of human perception.
However, common distance metrics such as MSE fall short as an effective PM.
In detail, for common metrics, high similarity scores ensure perceptual equivalence, but low scores do not guarantee perceptual differences.
Consider Figure \ref{example_for_metric_shortcoming}, where seemingly indistinguishable images to human eyes are deemed significantly different by MSE.
Consequently, relying on MSE risks pseudo-privacy protection.
This necessitates a PM that better reflects human perceptual thresholds for privacy leakage.

\subsubsection{Metric Design}
An ideal PM should satisfy the monotonicity principle: increasing $PM(x^*,x)$ corresponds to decreasing privacy information in $x^*$.
One intuitive idea for designing such a PM is to generate a reference sample devoid of any private information associated with $x$.
This reference sample acts as a benchmark to evaluate the amount of privacy information retained in $x^*$.
The closer $x^*$ is to the reference sample, the less private information from $x$ will be embedded in $x^*$.
However, a key challenge arises: how to obtain a suitable reference sample?

At first glance, one might consider utilizing random noises\footnote{Random noises are considered as non-information samples~\cite{info_theory}.} as the reference sample.
Though technically feasible, it produces a bottleneck for the performance of \sysname by restricting $x^*$ solely to a limited trajectory from $x$ to that specific reference, thus bypassing other possibly more informative regions of noises.
To address the problem, we define the distance of $x^*$ to the uniform distribution as our PM.
Unlike specific random noises, the uniform distribution offers a more flexible and adaptive benchmark.

Let $x^*$ be distributed according to Dirac delta distribution $p(x)$ that concentrates mass at $x^*$.
To measure the distance between $x^*$ and uniform noise distribution $q(x)$, we employ the JS divergence: $\frac{1}{2} \int p(x) log \frac{2p(x)}{p(x)+q(x)} + q(x) log \frac{2q(x)}{p(x)+q(x)} dx$.
Since directly evaluating JS divergence is inhibited by the high dimensionality of $x$, we reformulate JS divergence as follows (see Appendix \ref{proof_of_privacy_metric} for details):
\begin{equation}
\label{privacy_obj}
    \underset{D,D(x) \in [0,1]}{\max} \mathbb{E}_{x \sim p(x)} [log (1-D(x))] +  \mathbb{E}_{x \sim q(x)} [log D(x)].
\end{equation}

\textbf{Practical Implementation.}
Realizing Equation \ref{privacy_obj} entails constructing a new $D$ for each $x$ to estimate the JS divergence.
Seeking efficiency, we search for a single $D$ for all images.
We employ a neural network, dubbed the evaluation network, to serve this role, as DNNs are capable of approximating any function.
The goal of Equation \ref{privacy_obj} is to encourage $D$ to output 1 for random noises and 0 for natural samples.
Intuitively, the evaluation network in fact quantifies the proportion of noise within $x^*$.
This inspires us to create training data-label pairs for the network in an interpolate manner, \ie, $( (1-r) \cdot t_1 + r \cdot t_2, r), r \sim U(0,1), t_1 \sim p, t_2 \sim q$, and these pairs are supplied into the network as supervised signals for training.
Doing so enjoys that the monotonicity principle can be explicitly instilled into the evaluation network.

\revise{We train the evaluation network using TinyImageNet dataset~\cite{tinyimagenet} due to its wide range of common images (more training details can be found in Appendix \ref{leaving_detailed_settings}).
Figure \ref{evaluation_network_nums1} shows the network's ability to predict noise ratios across mixed samples of random noise and natural images, showcasing strong cross-dataset generalization capabilities.
Notably, it even delivers outstanding performance on significantly different datasets like SVHN (a dataset composed of digit images) \cite{svhn} and medical datasets including Path and Derma, further solidifying its wide-ranging applicability.
We highlight that this network can be trained using publicly available, non-sensitive datasets like TinyImageNet, preventing privacy leaks.
The abundance of public, non-private datasets makes it feasible for clients to have an evaluation network.
In practical scenarios, clients can obtain the trained evaluation network from a trusted third party at no cost or train their own evaluation network using publicly available data.
Moreover, the model has a compact size of approximately 1MB and requires less than 10 minutes of training time on a single GTX A10 GPU, making training cost manageable.
In this paper, all clients share this trained evaluation network by default\footnote{We observe comparable defense effectiveness when clients train individual networks using public data.}.}

\subsection{Solving Optimization Task (Equation \ref{optim_task})}
\label{solving_opt_problem}

\begin{figure}[!t]
    \centering
    \subfigure[]{
        \includegraphics[width=0.15\linewidth]{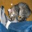}}
    \subfigure[]{
        \includegraphics[width=0.15\linewidth]{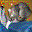}}
    \subfigure[]{
        \includegraphics[width=0.15\linewidth]{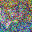}}
    \subfigure[]{
        \includegraphics[width=0.15\linewidth]{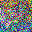}}
    \caption{We optimize the above image (a) and (c) for 10 iterations with learning rate of 1.
    The above image (c) represents a mixture of the original image and random noises, with $\alpha=0.5$. 
    The above image (b) and (d) denote the optimized images resulting from the above image (a) and (c), respectively. 
    Without noise-blended initialization, the adversarial vulnerability of DNNs results in generated images that maintain similar semantic information to the original image. }
    \label{opt_task_adv}
\end{figure}

Upon establishing the specific formulations for both $UM$ and $PM$, the remaining challenge lies in resolving Equation \ref{optim_task}.
A straightforward solution is to harness gradient descent algorithm~\cite{convex_opt}, widely recognized for its efficacy in solving optimization tasks that feature primarily convex components, such as $UM$.
However, the algorithm encounters obstacles when confronted with highly non-linear and non-convex neural networks like PM.

To show this, we initialize $x^*$ as $x$ and then maximize $PM(x,x^*)$, \ie, with the hope of resultant $x^*$ being noisy.
As shown in Figure \ref{opt_task_adv}(b), the resultant $x^*$ deviates only slightly from $x$ (Figure \ref{opt_task_adv}(a)), yet the model confidently identifies it as a noisy sample.
This phenomenon and $x^*$ are known as adversarial vulnerability and adversarial examples~\cite{FGSM,CW}.

To address the above problem, we propose noise-blended initialization which mixes $x^*$ with random noises, \ie, $x^*=(1-\alpha) x+ \alpha v, \ v \sim q(x), \alpha \in [0,1]$.
$\alpha$ is a blend factor.
A high value of $\alpha$ positions $x^*$ closer to areas dense with noises and ensures that the evaluation network will maintain $x^*$ as a noisy sample during the optimization process.
Figure \ref{opt_task_adv}(c) exemplifies the noise-blended initialization with $\alpha=0.5$.
The resultant $x^*$ is devoid of semantic information regarding $x$, showing the effectiveness of the noise-blended initialization.

\section{Theoretical Analysis}
\label{framework}

\subsection{Performance Maintenance}
\label{subsec_convergence}

\begin{assumption}
\label{assumpt1}
$\mathcal{L}(F_{\theta}(x),y)$ is $L$-smooth: $\forall \theta_1, \theta_2,$ it holds that $\mathcal{L}(F_{\theta_1}(x),y) \leq \mathcal{L}(F_{\theta_2}(x),y) + \nabla_{\theta_2} \mathcal{L}(F_{\theta_2}(x),y)^T (\theta_1-\theta_2) + \frac{L}{2}||\theta_1 - \theta_2||_2^2$.
\end{assumption}

\begin{assumption}
\label{assumpt2}
$\mathcal{L}(F_{\theta}(x),y)$ is $u$-strongly convex: $\forall \theta_1, \theta_2,$ there is $\mathcal{L}(F_{\theta_1}(x),y) \geq \mathcal{L}(F_{\theta_2}(x),y) + \nabla_{\theta_2} \mathcal{L}(F_{\theta_2}(x),y)^T (\theta_1-\theta_2) + \frac{u}{2}||\theta_1 - \theta_2||_2^2$.
\end{assumption}

\begin{assumption}
\label{assumpt3}
Let $x_i, y_i$ be uniformly random samples drawn from the local dataset $D_i$ of the $i$-th client. The variance of stochastic gradients is bounded: $\mathbb{E} || \nabla_{\theta} \mathcal{L}(F_{\theta}(x_i),y_i) - g_{i}^{full} ||_2^2 \leq \sigma_i^2, i=1,2,\cdots,K.$
Wherein, $g_{i}^{full}$ represents the gradients of model parameters $\nabla_{\theta} \mathcal{L}(F_{\theta}(\cdot),\cdot)$ computed across the complete local dataset $D_i$.
\end{assumption}

\begin{assumption}
\label{assumpt4}
There exists a real number $G$ and the expected squared norm of stochastic gradients is uniformly bounded by $G$: $\mathbb{E} ||\nabla_{\theta} \mathcal{L}(F_{\theta}(x_i),y_i)||_2^2 \leq G^2, i=1,2,\cdots,K$.
\end{assumption}

\begin{thm}
\label{convergence_thm}
Let Assumption \ref{assumpt1}, \ref{assumpt2}, \ref{assumpt3}, and \ref{assumpt4} hold.
Let $x^*$ be the optimal solution of Equation \ref{optim_task}.
Suppose the gradient distance associated with $x$ and $x^*$ is bounded: $|| \nabla_{\theta} \mathcal{L}(F_{\theta}(x),y) - \nabla_{\theta} \mathcal{L}(F_{\theta}(x^*),y)||_2^2 \leq \epsilon^2$.
Let $\theta^*$ represent the global optimal solution and $\theta_i^*$ ($i=1,2,\cdots, K$) denote the local optimal solution for $i$-th client.
Choose $\kappa=\frac{L}{u}, \gamma=\max \{ 8\kappa, 1 \}, \eta = \frac{2}{u(\gamma+t)}$.
Let $\Gamma = \mathbb{E}[\mathcal{L}(F_{\theta^*}(\cdot),\cdot)] -  \frac{1}{K} \sum_{i=1}^{K} \mathbb{E}[\mathcal{L}(F_{\theta_i^*}(\cdot),\cdot)]$.
There is:
\begin{equation}
\begin{split}
    &\mathbb{E}[\mathcal{L}(F_{\theta}(\cdot),\cdot)] - \mathcal{L}(F_{\theta^*}(\cdot),\cdot) \\
    & \leq \frac{2 \kappa}{\gamma + N}(\frac{Q+C}{u} + \frac{u \gamma}{2} \mathbb{E} ||\theta_{initial} - \theta^*||_2^2 ), \nonumber
\end{split}
\end{equation}
where $Q = \frac{1}{K^2}\sum_{i=1}^K ( \epsilon^2 + \sigma_k^2) + 6L \Gamma, \  C = \frac{4}{K} ( \epsilon^2 + G^2)$, and $\theta_{initial}$ is initial parameters of the globally-shared model.
\end{thm}
This subsection examines how \sysname contributes to the convergence of the global model.
We here demonstrate Theorem \ref{convergence_thm} by making Assumption \ref{assumpt1}-\ref{assumpt4}.
See \ref{proof_of_convergence} for the proof of Theorem \ref{convergence_thm}.
\revise{Assumption \ref{assumpt1} posits that DNNs' loss landscape does not exhibit abrupt curvature changes, which is satisfied by common DNNs and empirically validated \cite{lipschitz_est2}.
Assumption \ref{assumpt2} follows the technical conventions.
Notably, existing theoretical work \cite{milne2019piecewise} demonstrated that DNNs exhibit strong convexity near local optima.
From initialization perspective, DNNs' parameters reside within a basin of attraction and converge to the corresponding optimal point (though some methods \cite{deng2020contour} explore transitions between basins).
Assumptions \ref{assumpt3} and \ref{assumpt4} are intuitively valid.
In practice, it is almost impossible for the gradient from certain data points to be infinitely far from the full dataset's gradient or infinitely large (Assumptions \ref{assumpt3} and \ref{assumpt4}).
As a result, Theorem \ref{convergence_thm} enjoys good applicability over common DNNs.}

Theorem \ref{convergence_thm} delineates an upper bound on the gap between the optimal solution and the model trained with \sysname after sufficient rounds.
As the gradient distance between robust and client data decreases, this upper bound also decreases, implying better performance of the global model.
This highlights the effectiveness of minimizing the gradient distance between robust data and clients' data in maintaining performance.
The inclusion of the gradient distance term ($UM$) in Equation \ref{optim_task} guarantees the boundedness of $|| \nabla_{\theta} \mathcal{L}(F_{\theta}(x),y) - \nabla_{\theta} \mathcal{L}(F_{\theta}(x^*),y)||$.
In practice, we modify the gradients of robust data slightly.

If the difference between the gradients of robust data and clients' data exceeds $\epsilon$, we search for the closest gradients within the $\epsilon$-constraint and use them as the final uploaded gradients (see Appendix \ref{projection_algorithm} for more details).
Doing so enables us to evaluate the utility-privacy trade-off of \sysname by varying values of $\epsilon$\footnote{Although adjusting the value of $\beta$ is also an alternative way, we find that tuning $\epsilon$ is more convenient.}.

Moreover, FL commonly involves two scenarios: when the client data distributions are identical (IID) and non-identical (Non-IID).
Our theoretical analysis does not build upon homogeneity across client datasets, rendering Theorem \ref{convergence_thm} applicable to both scenarios.

\subsection{Privacy Protection}
\label{theory_privacy_part}

GLAs vary widely in formulation but share a common goal: reconstructing training data from gradients via inverse mappings $\hat{f}^{-1}(\cdot)$.
To enable broad privacy analysis, we abstract GLAs into a unified framework where $f(x) = g$ represents the forward gradient mapping, and $\hat{f}^{-1}(g)$ approximates the inverse mapping with bounded error $\sigma$ (i.e., $\text{dist}(\hat{f}^{-1}(\cdot),f^{-1}(\cdot)) \leq \sigma$).
$\text{dist}(\cdot,\cdot)$ is unspecified distance metric and $\sigma$ is a small positive number.
This abstraction accommodates diverse attack variants (Examples \ref{attack_example2}-\ref{attack_example3}) by focusing on their shared objective of minimizing $\text{dist}(\hat{f}^{-1}(g),x)$.

\begin{example}
\label{attack_example2}
\textit{
    (Inverting Gradients~\cite{invert_grad}).
    Similar to DLG, \citet{invert_grad} also solves a gradient matching problem for data reconstruction.
    But $L_2$-norm distance is replaced by cosine similarity and a total variation term used to regularize reconstructed data is introduced:
    $ \hat{x} = \hat{f}^{-1}(g) = \mathop{\arg\min}_{\hat{x}}  1-\frac{<\nabla_{\theta} F_{\theta}(\hat{x}), g>}{|| \nabla_{\theta} F_{\theta}(\hat{x} ||_2 || g ||_2} + TV(x) $.
}
\end{example}


\begin{example}
\label{attack_example3}
\textit{
    (Generative Gradient Leakage (GGL)~\cite{gga}).
    GGL optimizes a latent input vector of a generative model to search data whose gradients align well with uploaded gradients.
    Moreover, GGA adds a KL divergence term to avoid too much deviation between the latent vector and the generator's latent distribution:
    $ \hat{x} = \hat{f}^{-1}(g) = G(z)$, where $z = \mathop{\arg\min}_{x}  || \nabla_{\theta} F_{\theta}(G(z)) - g ||_2^2 + KL(z||q)$. $q$ is Gaussian distribution with a mean of 0 and a standard deviation of 1.
}
\end{example}

When employing \sysname, clients upload gradients associated with $x^*$, denoted by $g^* = f(x^*)$.
The server exploits $g^*$ to recover data, \ie, $\hat{f}^{-1}(g^*)$.
We now consider the distance of $x$ and $\hat{f}^{-1}(g^*)$:
\begin{equation}
\label{lower_bound_1}
\begin{split}
        &\text{dist}(\hat{f}^{-1}(g^*),x) = \text{dist}(\hat{f}^{-1}(g^*),f^{-1}(g))
        \\
        & \geq | \ \text{dist}(f^{-1}(g^*),f^{-1}(g)) - \text{dist}(\hat{f}^{-1}(g^*),f^{-1}(g^*)) \ |
        \\
        &= | \ \text{dist}(x^*,x) -\text{dist}(\hat{f}^{-1}(g^*),f^{-1}(g^*)) \  |.
\end{split}
\end{equation}
Equation \ref{lower_bound_1} establishes a lower bound of $\text{dist}(\hat{f}^{-1}(g^*),x)$, which hinges upon $x^*$, $x$, and $\sigma$.
In practice, $x^*$ obtained by Equation \ref{optim_task} are often noisy and significantly differ from $x$, \ie, $\text{dist}(x^*,x) > \sigma$ (Appendix \ref{appendix_privacy_theoy_exp}).
When the value of $\sigma$ is small, the lower bound is simplified as follows:
\begin{equation}
\label{lower_bound_2}
\text{dist}(\hat{f}^{-1}(g^*),x) \geq  \text{dist}(x^*,x) - \sigma.
\end{equation}
Equation \ref{lower_bound_2} suggests that a smaller $\sigma$ results in higher privacy lower bound, \ie, stronger privacy protection capabilities.
On the other hand, considering a large $\sigma$ may be a little trivial, because a large $\sigma$ leads to a significant discrepancy between the recovered data and clients' data.

To delve deeper, we adopt two specific metrics, including Euclidean distance and our evaluation network.
The use of Euclidean distance enables Equation \ref{lower_bound_2} to be restated:
\begin{equation}
||\hat{f}^{-1}(g^*) - x||_2^2 \geq ||x^* - x||_2^2 - \sigma.
\end{equation}
The distance for the evaluation network is defined as the absolute difference in noise ratios between the two data\footnote{See Appendix \ref{appendix_proof_evaluation_network_distance_metric} for proof of the definition qualified as distance metric.}.
Given that $x$ is natural data, $\text{dist}(\hat{f}^{-1}(g^*),x)$ and $\text{dist}(x^*,x)$ reflect the amount of random noises present within $\hat{f}^{-1}(g^*)$ and $x^*$.
Thus, Equation \ref{lower_bound_2} with the evaluation network is translated as follows:
\begin{equation}
D(\hat{f}^{-1}(g^*)) \geq  D(x^*) - \sigma.
\end{equation}
The above equation indicates that, if $x^*$ carries a substantial noise fraction, the data recovered by attackers, too, will be significantly distorted by noises.

\subsection{Time Complexity}

\begin{table}[]
\caption{Time complexity comparison. $\mathcal{N}, \mathcal{N}_{last}, S_1, S_2, I$ denote the total number of model parameters, the total number of model parameters in the last fully connected layer, the time required for one forward-backward propagation of the globally-shared model and the evaluation network, the iteration number for solving Equation \ref{optim_task}, respectively.}
\label{theor_time_comp}
\centering
\footnotesize
\begin{tabular}{@{}ccccc@{}}
\toprule
DP & GQ & Pruning & Soetria & Ours \\ \midrule
$\mathcal{O}(\mathcal{N})$ & $\mathcal{O}(\mathcal{N})$  & $\mathcal{O}(\mathcal{N} \log (\mathcal{N}) )$ & $\mathcal{O}(\mathcal{N}_{last} \cdot  S_1)$ & $\mathcal{O}( \iota \cdot (2 \cdot S_1 + S_2 ) )$ \\ \bottomrule
\end{tabular}
\end{table}

Before analyzing time complexity, it is essential to recognize that achieving excellence in all aspects of utility, privacy, and time is almost impossible.
Generally, improving the utility-privacy balance demands additional time investment.
Consider DP, which equally perturbs all gradient elements without considering their privacy information.
In contrast, Soteria estimates the privacy information of each gradient element and selectively prunes those with the highest privacy information, thereby improving the trade-off but at the cost of extra time incurred by the estimation process.

DP, GQ, and gradient pruning, none of which necessitate extra forward-backward propagation, are discussed together.
DP generates noises for each gradient element, GQ quantizes each gradient element, and gradient pruning sorts and then removes part of the gradient elements.
Assuming a model with $\mathcal{N}$ total parameters, the time complexity of both DP and GQ scales linearly with the number of parameters, $\mathcal{O}(\mathcal{N})$.
For gradient pruning, the dominant factor is the sorting time complexity, \ie, $\mathcal{O}(\mathcal{N} \log (\mathcal{N}))$.
Soteria estimates privacy information for each gradient element in the fully connected layer\footnote{Typically, a fully connected layer comprises around 1024 neurons.}.
Given that the total count of parameters contained within the layer is to be denoted as $\mathcal{N}_{last}$, and once forward-backward propagation of the global model holds a time complexity of $\mathcal{O}(S_1)$.
Soteria iteratively performs forward-backward propagations a specific number of times, corresponding to the number of neurons in the layer.
Consequently, the overall time complexity of Soteria can be expressed as $\mathcal{O}(\mathcal{N}_{last} \cdot  S_1)$.

The time complexity of \sysname centers around solving Equation \ref{optim_task}.
Each iteration of Equation \ref{optim_task} requires twice forward-backward propagations of the global model, along with one forward-backward propagation of the evaluation network.
Hence, the overall time complexity is the sum of these operations.
Assuming the time complexity of a single forward-backward propagation of the evaluation network is $\mathcal{O}(S_2)$ and the quantity of iterations required to solve Equation \ref{optim_task} is $\iota$, the time complexity of \sysname is $\mathcal{O}( \iota \cdot (2 \cdot S_1 + S_2 ) )$.
Table \ref{theor_time_comp} summarizes the time complexities of these defenses.

\subsection{Security Discussion}
\label{security_discussion}

We here conduct a security analysis of \sysname, \ie, the resilience of \sysname against adaptive attacks.
Let us consider a white-box scenario where attackers have complete access to any knowledge of the deployed defense mechanisms, excluding $x$ and $g$.
According to our theoretical analysis, attackers can reconstruct $x^*$ via clients' uploaded gradients $g^*$.
Our method centers on Equation \ref{optim_task}, with attackers attempting to reverse-engineer the ground-truth gradients associated with $x$ using Equation \ref{optim_task}.
Consequently, Equation \ref{optim_task} boils down to $||g^* - g||_2^2 = const$, where all constant terms are absorbed into $const$, and we omit gradient weighting factors for notation simplicity.
Owing to the high dimensionality of the gradients, this equation stands as inherently under-determined, unless the gradient dimension is 1, which is nearly impossible in real-world scenarios.
Further discussions and validations can be found in Appendix \ref{further_security_analysis}.

\section{Evaluation Setup}
\label{evaluation_setup}

\begin{figure*}[!h]
	\centering
		\includegraphics[width=0.6\linewidth]{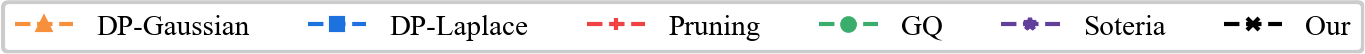}
        \\
	\subfigure[iDLG]{
		\includegraphics[width=0.23\linewidth]{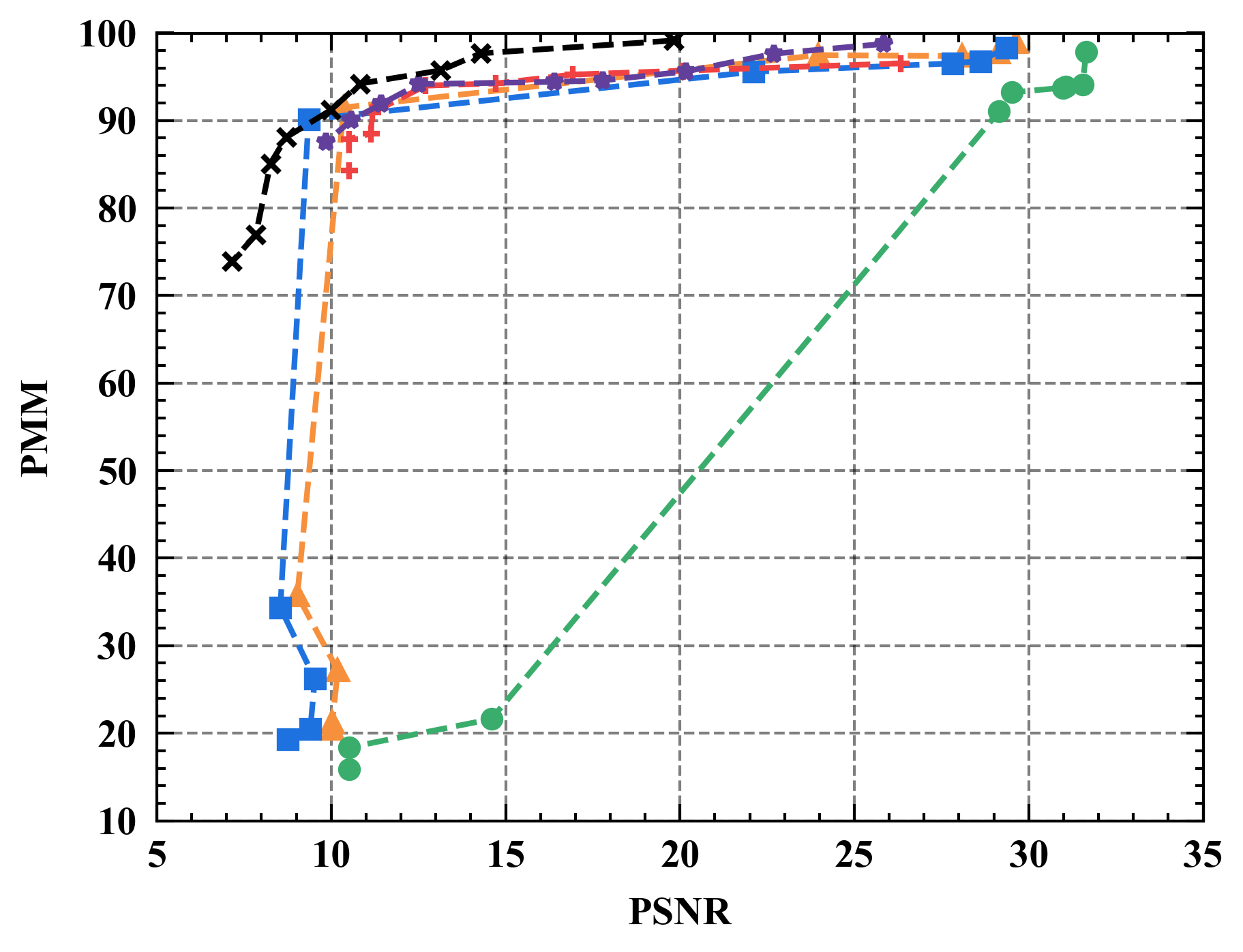}}
	\subfigure[GradInversion]{
		\includegraphics[width=0.23\linewidth]{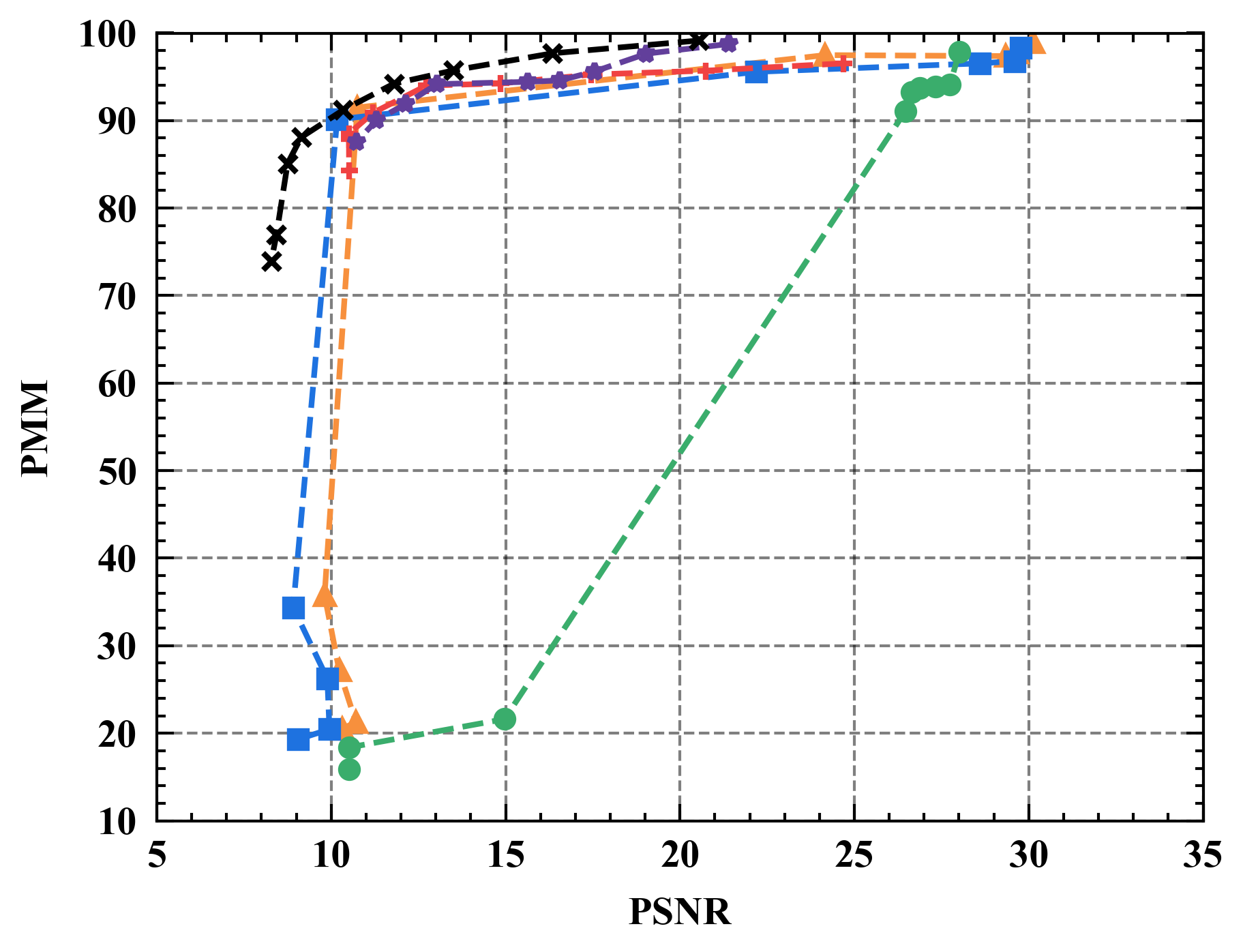}}
	\subfigure[InvertingGrad]{
		\includegraphics[width=0.23\linewidth]{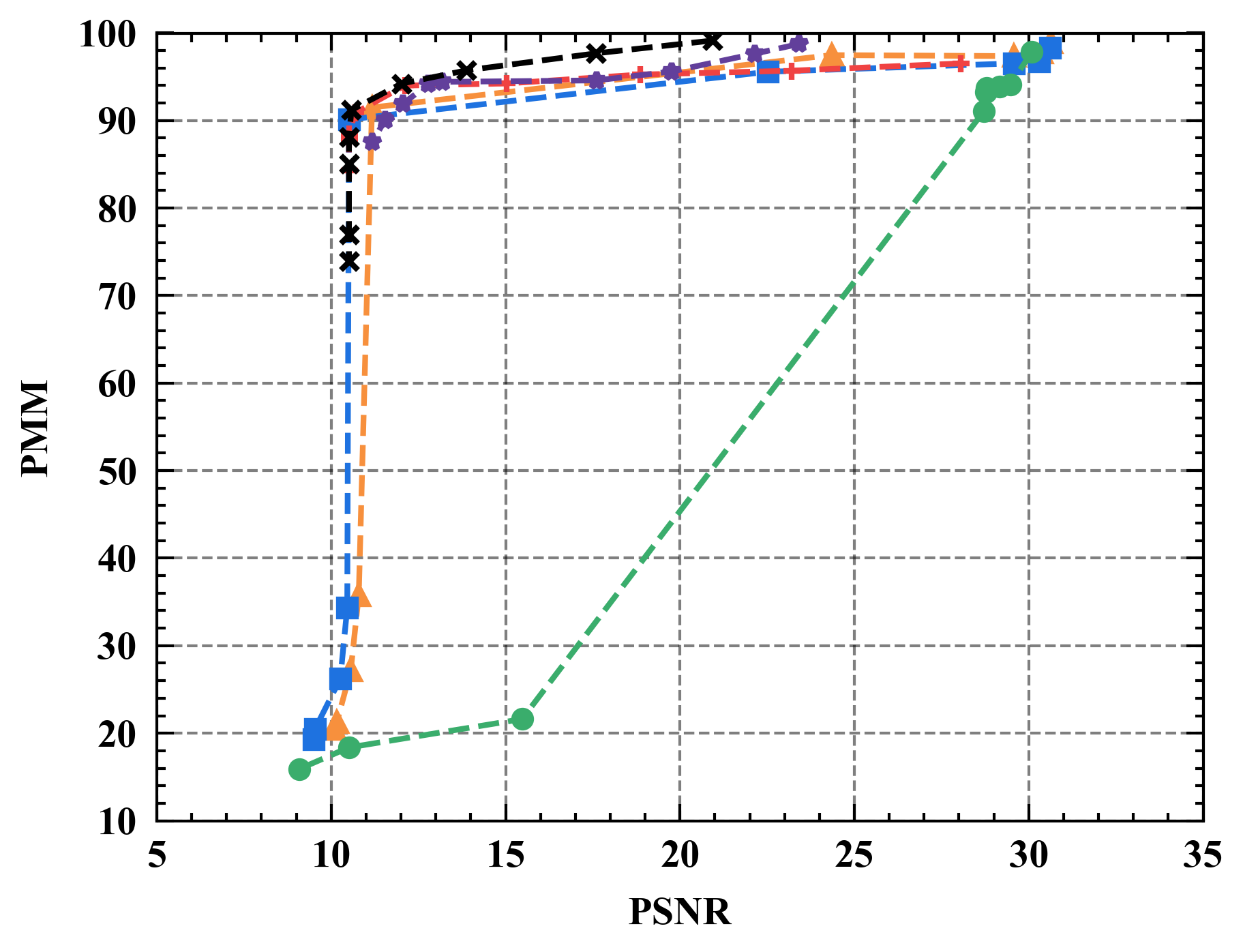}}
  	\subfigure[GGL]{
		\includegraphics[width=0.23\linewidth]{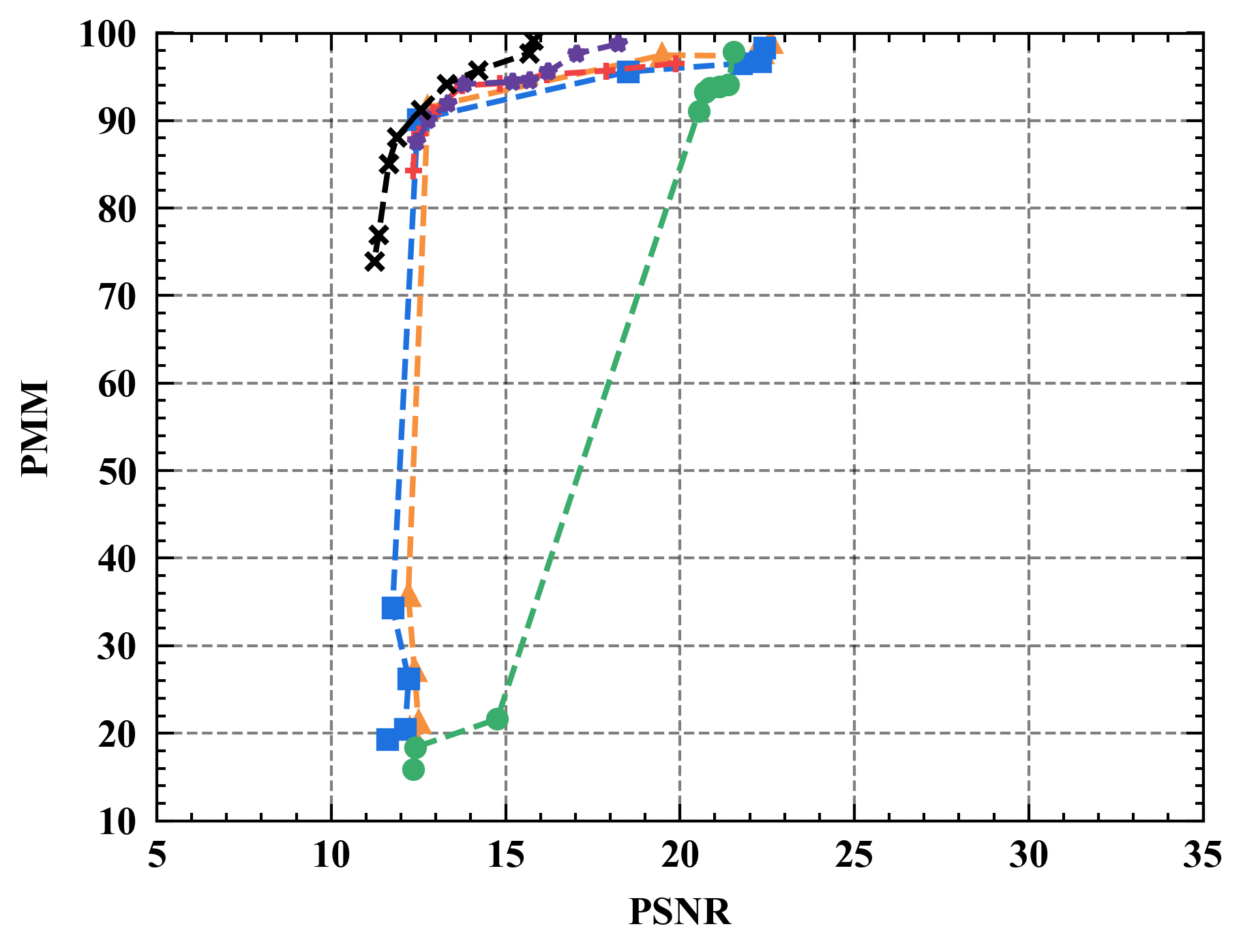}}
	\caption{The performance (PMM-PSNR curves) of defenses against four state-of-the-art attacks in LeNet trained with CIFAR10.}
	\label{res_lenet_cifar10_psnr} 
\end{figure*}

\begin{figure}[t]
    \centering
    \includegraphics[width=0.4\textwidth]{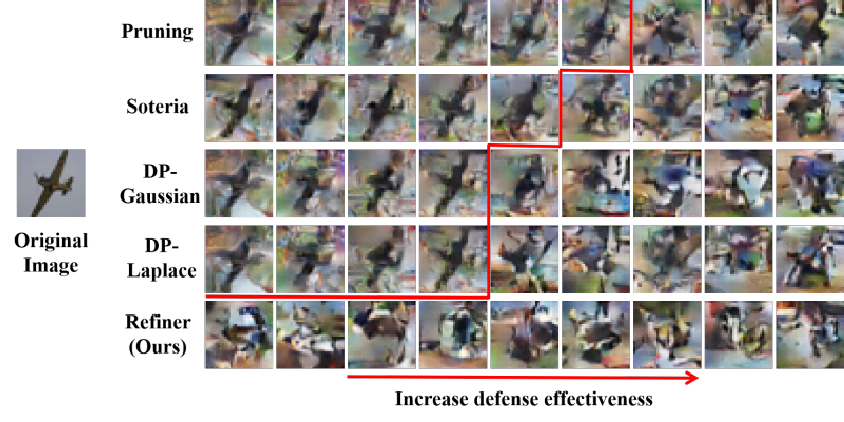}
    \caption{A visualization of five defenses over varying hyperparameters against GGL. The images above the red broken lines contain privacy information of the original images.}
    \label{recovered_image_visualization}
\end{figure}

\begin{figure} [t]
	\centering
	\subfigure[LPIPS]{
		\includegraphics[width=0.48\linewidth]{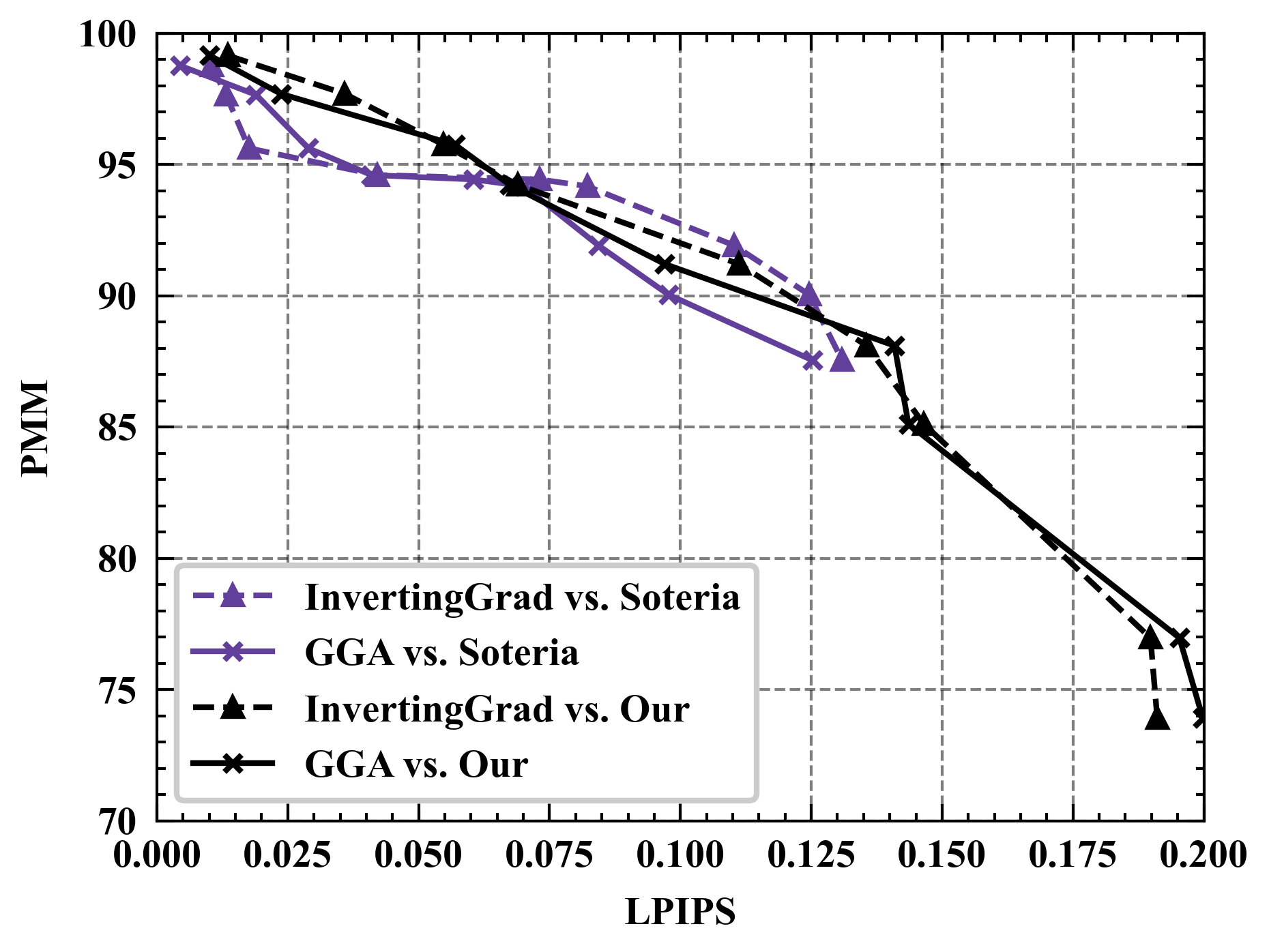}}
	\subfigure[Evaluation Network]{
		\includegraphics[width=0.48\linewidth]{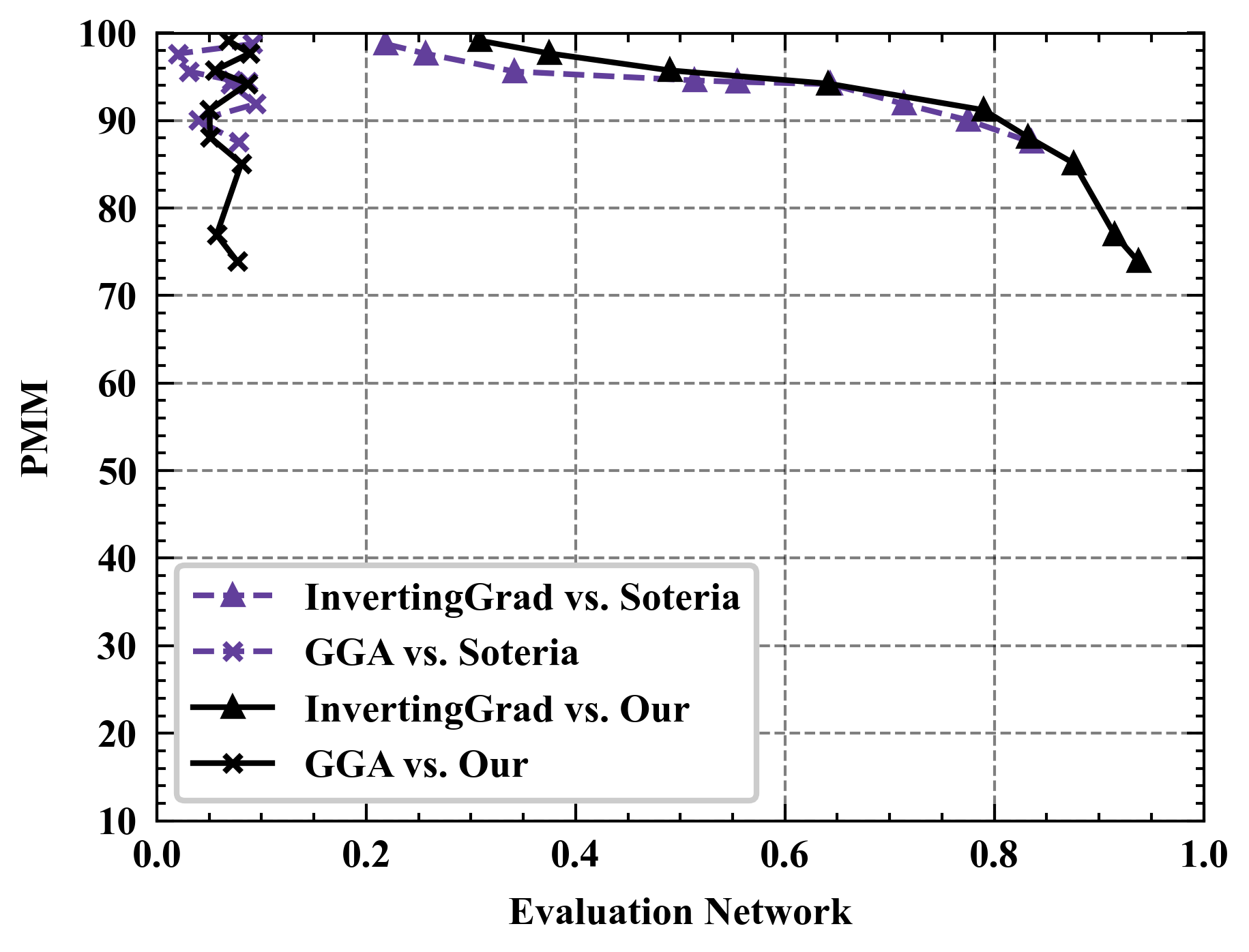}}
	\caption{The defense performance of Soteria and \sysname against InvertingGrad and GGL using LPIPS and the evaluation network to assess. We use LeNet and CIFAR10.}
	\label{exp_GGA} 
\end{figure}

\begin{figure*} [t]
	\centering
		\includegraphics[width=0.6\linewidth]{images/attack_res/legend.png}
        \\
	\subfigure[SSIM]{
		\includegraphics[width=0.24\linewidth]{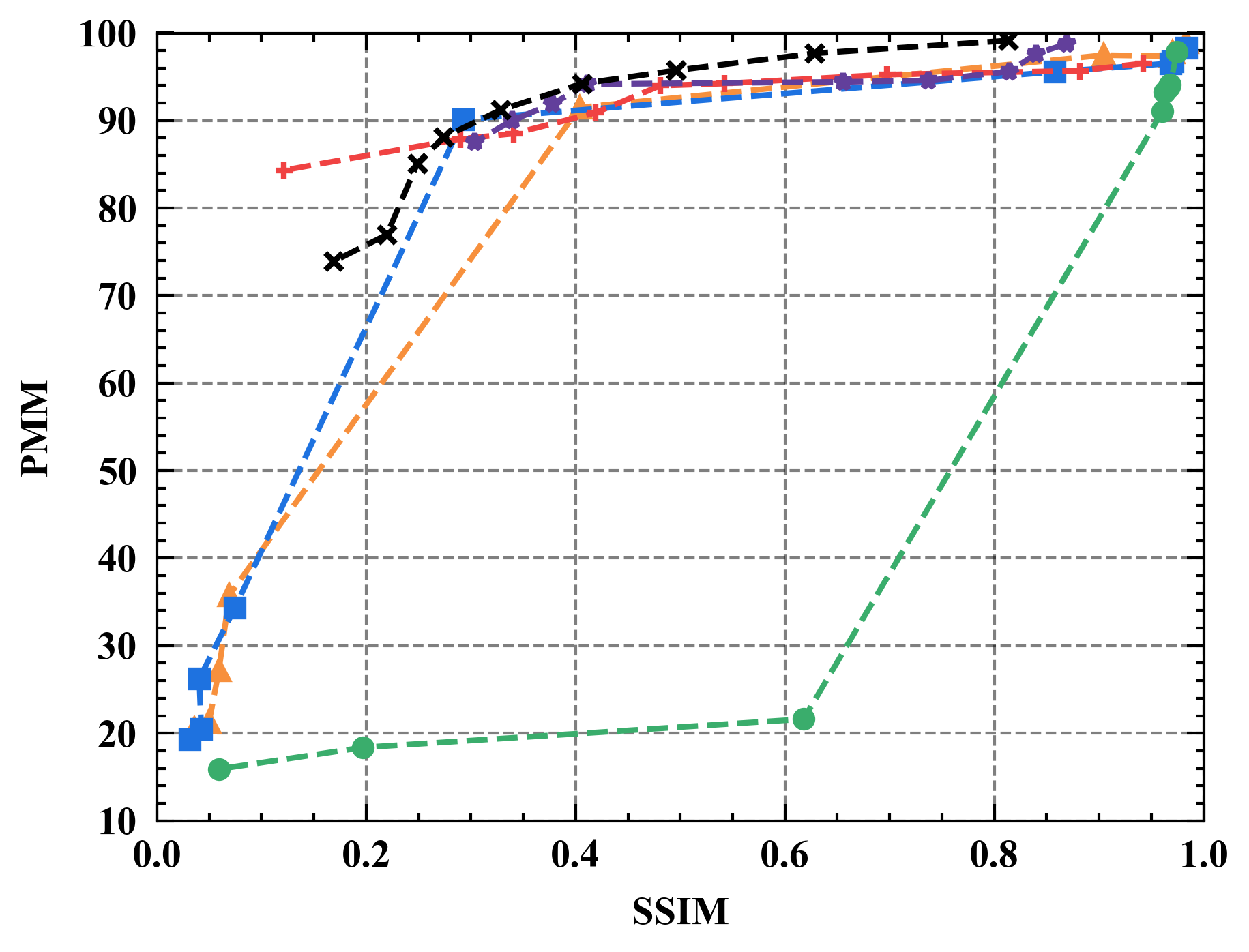}}
	\subfigure[LPIPS]{
		\includegraphics[width=0.24\linewidth]{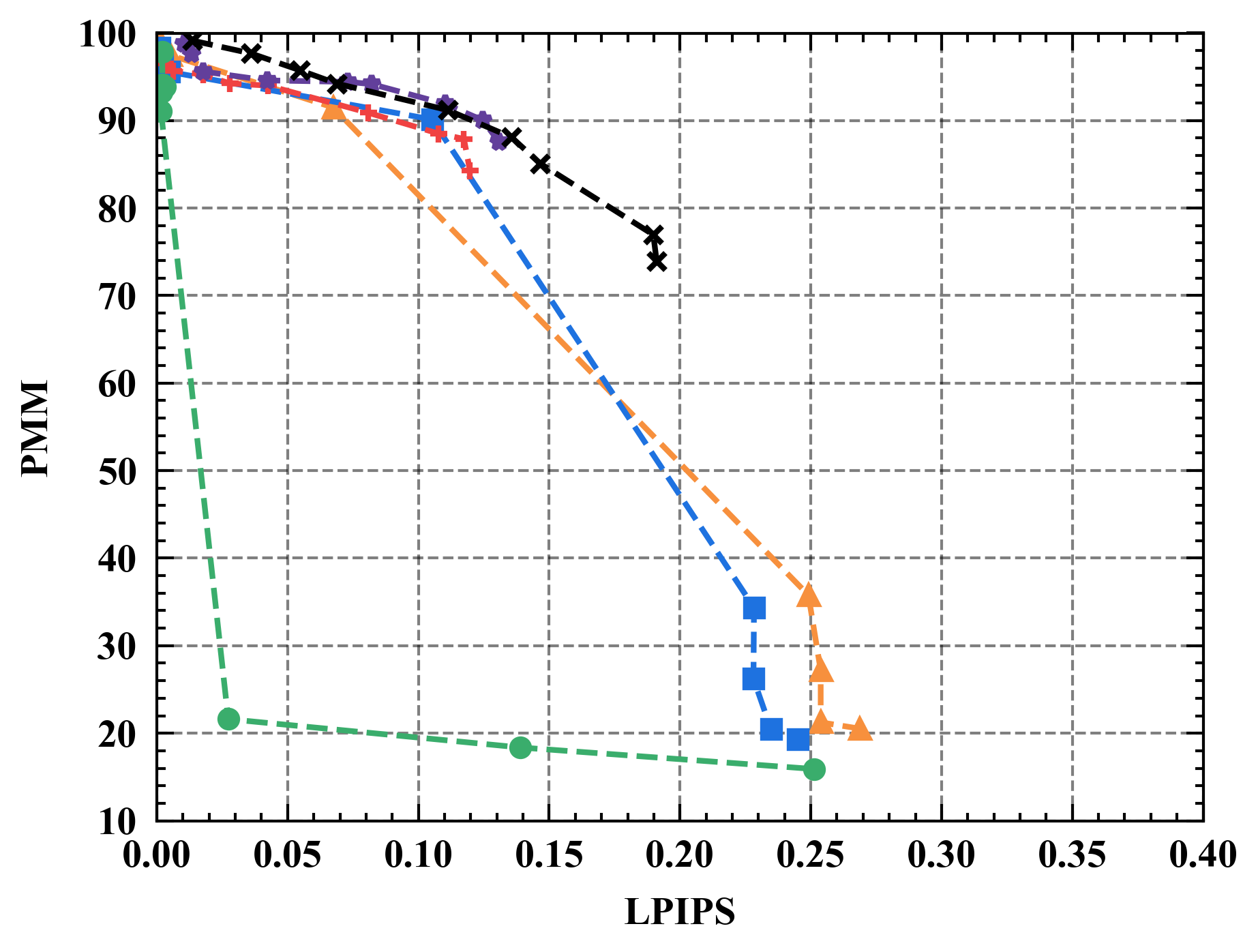}}
	\subfigure[Evaluation Network]{
		\includegraphics[width=0.24\linewidth]{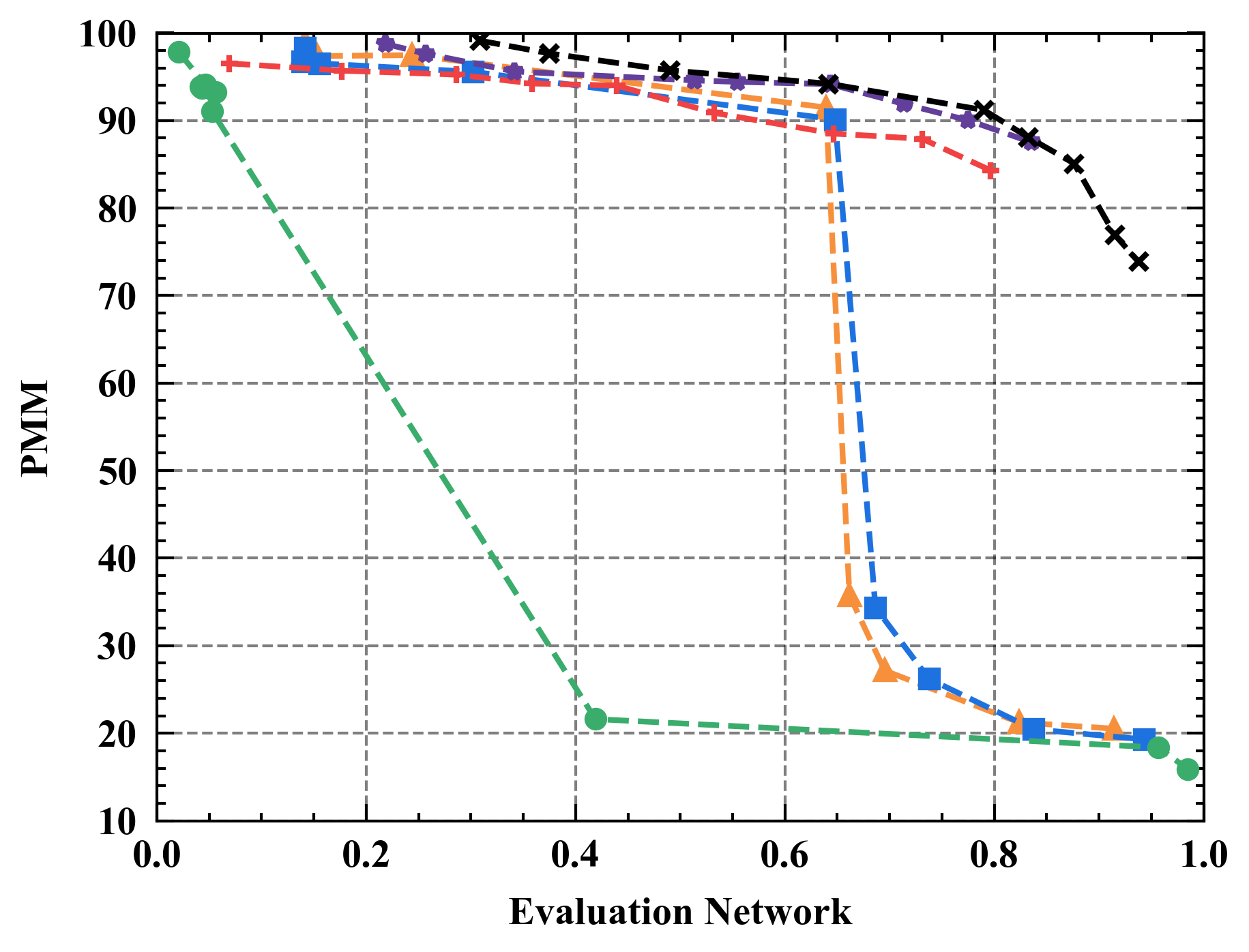}}
	\caption{The performance of defenses against InvertingGrad in LeNet trained with CIFAR10 using different metrics to assess.}
	\label{exp_diff_metric} 
\end{figure*}

\begin{figure*}[!t]
	\centering
		\includegraphics[width=0.6\linewidth]{images/attack_res/legend.png}
        \\
	\subfigure[ResNet10]{
		\includegraphics[width=0.24\linewidth]{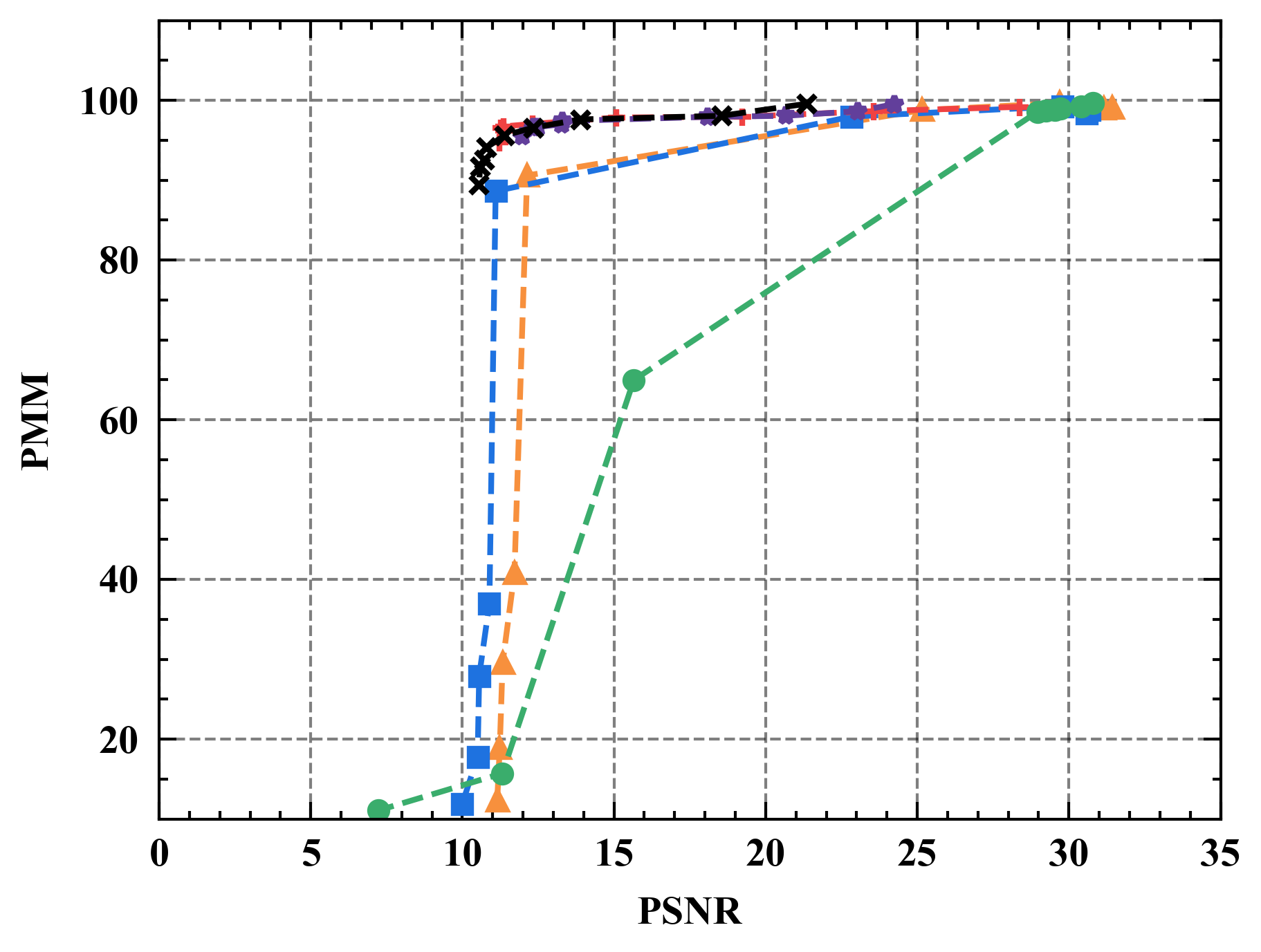}}
	\subfigure[ResNet18]{
		\includegraphics[width=0.24\linewidth]{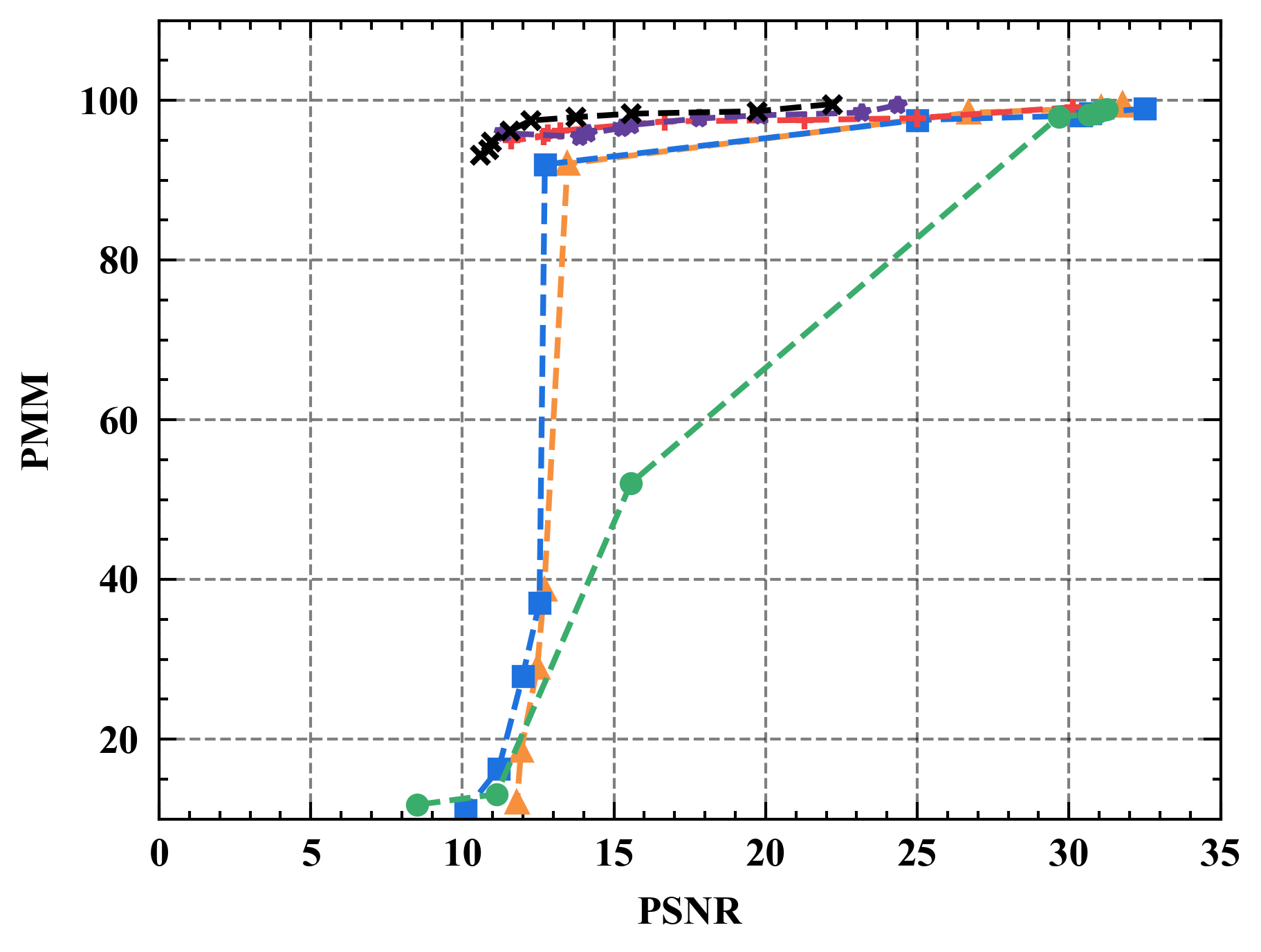}}
  	\subfigure[ViT]{
		\includegraphics[width=0.24\linewidth]{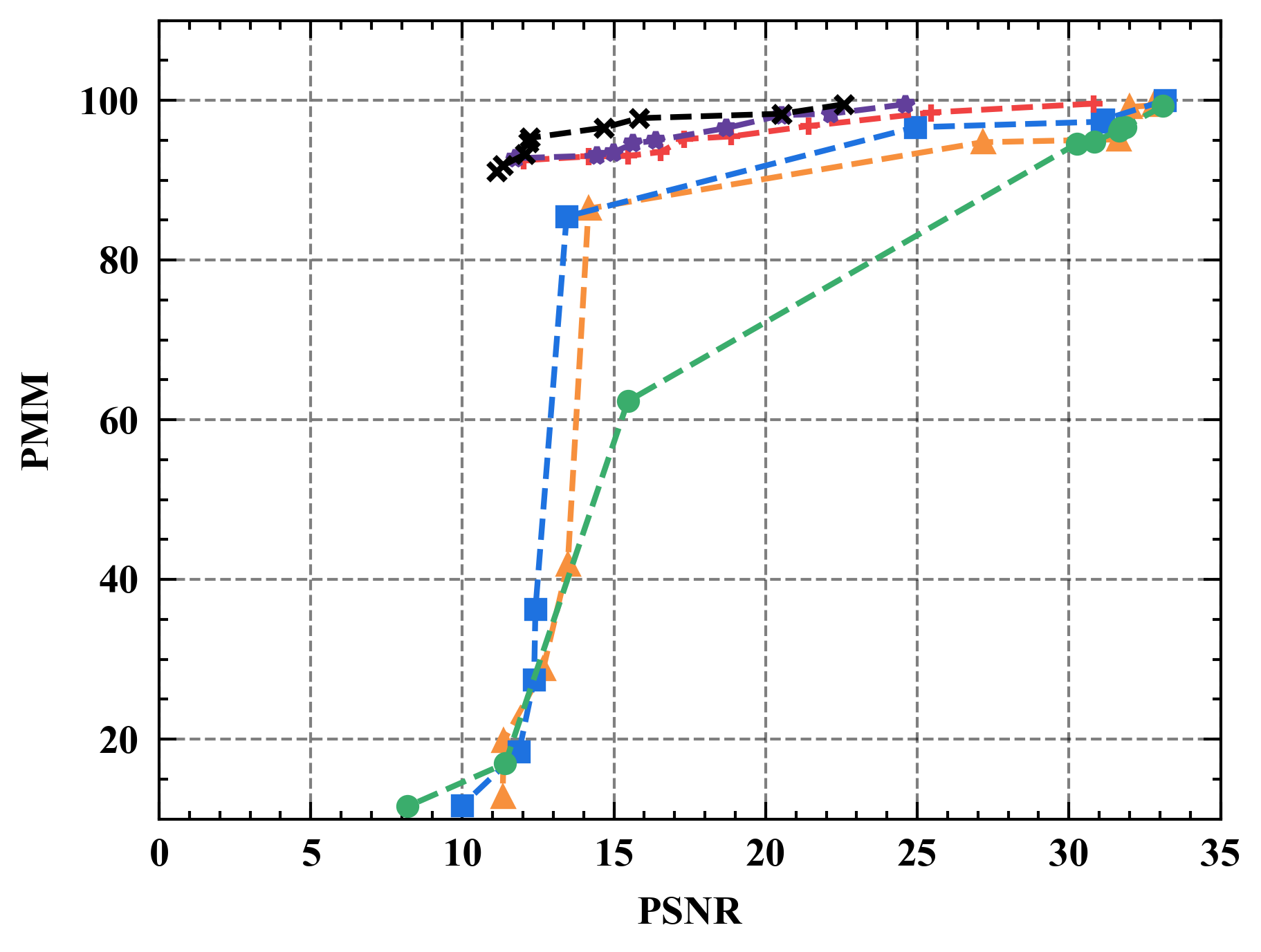}}
	\caption{The performance of various defenses against InvertingGrad in ResNet10, ResNet18 and ViT trained with CIFAR10.}
	\label{exp_diff_model} 
\end{figure*}

\subsection{Attack}
Four state-of-the-art attacks are used to test the performance of \sysname, including iDLG \cite{igla}, GradInversion \cite{grad_inversion}, InvertingGrad \cite{invert_grad} and GGL \cite{gga,grad_obfuscation}.
These attacks cover the primary types of GLAs currently explored.
The former three attacks address variants of Equation \ref{reconstruction_process} by employing varying optimizers, loss functions, etc.
GGL diverges from this path by leveraging latent vector alignments within a GAN model.
We implement these attacks at the start of training, as this stage is most vulnerable to GLAs \cite{bayes_attack_for_gla}.
For GGL, we use public codes \footnote{\url{https://raw.githubusercontent.com/pytorch/examples/master/dcgan/main.py}} to train a GAN.
See Appendix \ref{leaving_detailed_settings} for an overview of the attack distinctions and the hyperparameters employed.

\subsection{Competitors}
We compare \sysname to the following defenses: DP~\cite{dp}, GQ~\cite{fed_quant,grad_obfuscation}, gradient pruning~\cite{grad_pruning,gla}, and Soteria~\cite{soteria2}\footnote{We use Soteria proposed in \cite{soteria2} to compare, which is an improved version compared with the original Soteria~\cite{soteria}.}.
Each of these defenses involves a utility-privacy trade-off that is regulated by hyperparameters such as noise magnitude for DP, discretization level for GQ, and pruning rate for pruning and Soteria.
We vary these hyperparameters to obtain the utility-privacy trade-off curves for each defense. 
Specifically, for DP, we employ two commonly used kinds of noises, namely Gaussian (DP-Gaussian) and Laplace (DP-Laplace), with magnitude ranging from $10^{-6}$ to $10^{2}$ and $C$ of 1.
As for GQ, we consider discretization to 1, 2, 4, 8, 12, 16, 20, 24, and 28 bits.
The pruning ratio for gradient pruning and Soteria is chosen from the range ${0.1, 0.2, \cdots, 0.9}$.
Besides, we alter $\epsilon$ of \sysname over $\{0.01, 0.03, 0.05, 0.07, 0.1, 0.3, 0.5, 0.7, 0.9\}$ (see Section \ref{subsec_convergence}).
Unless otherwise specified, for \sysname, the default values of $\alpha$ (blend factor in noise-blended initialization), $\beta$ (balance factor in Equation \ref{optim_task}), $\iota$ (iterations for solving Equation \ref{optim_task}), and $\tau$ (decay factor in layer-wise weight) are set to 0.5, 1, 10, and 0.95.
\revise{These values are identified through preliminary experiments as effective in balancing privacy and utility, with our results demonstrating that \sysname's performance is not highly sensitive to hyperparameter choices.}

\subsection{Evaluation Metrics}

We assess defenses from three fundamental dimensions: performance maintenance, privacy protection, and time cost.

\textbf{Performance maintenance.}
We define the performance maintenance metric (PMM), which calculates the accuracy ratio achieved with defenses compared to the original accuracy without any defenses:
$
PMM = \frac{ defense\_acc}{original\_acc} \times 100\%,
$
where $original\_acc$ and $defense\_acc$ denote the accuracy of the trained global-shared model with and without defense over test sets of corresponding datasets\footnote{The original accuracies of LeNet are 84.12\%, 54.01\%, and 21.45\% on SVHN, CIFAR10, and CIFAR100. The original accuracies of ResNet10, ResNet18 and ViT are 83.25\%, 84.95\%, 87.52\% on CIFAR10.}.

\textbf{Privacy protection.}
Privacy protection measures the amount of privacy information recovered from the reconstructed images.
\revise{Given the lack of a universally perfect privacy assessment metric, we employ a diverse range of metrics, including PSNR~\cite{PSNR} ($\downarrow$), SSIM~\cite{SSIM} ($\downarrow$), LPIPS~\cite{LPIPS} ($\uparrow$), and the prediction noise ratio of the evaluation network ($\uparrow$), to achieve a more comprehensive assessment.}
Thresholds of 15 for PSNR, 0.5 for SSIM, 0.05 for LPIPS, and a noise ratio of 0.4 from the evaluation network represent the junctures at which human eyes struggle to discern any private information of the original data from recovered data.
Please see Appendix \ref{concrete_formula_assess_privacy} for a visual guide that illustrates how changes in these values affect the level of privacy exposure.

\textbf{Time cost.}
Time cost is also an important dimension for evaluating a defense. We record the time required for a single iteration of the model when employing defenses.

\subsection{Model, Dataset, and Attack Settings}

We select three widely-used models, LeNet, ResNet, and ViT~\cite{vision_transformer}, on three benchmark datasets SVHN, CIFAR10, and CIFAR100.
LeNet and ResNet are CNN-based models while ViT is a transformer-based model, covering mainstream model architectures.
We include ResNet10 and ResNet18 to study the effects of model scale.
Our focus is on a typical FL scenario involving 10 clients collaborating to train a global model, with a server.

Our evaluation also includes non-IID settings~\cite{non_iid_partition}.
Following~\cite{sample_method1}, the IID setting equitably and randomly segments training datasets among clients, while the non-IID setting introduces label imbalance, distributing data unevenly among the 10 clients~\cite{dirichlet_dist,non_iid_partition}.

\textbf{Training setting.}
The FL training process takes place over 8000 rounds, during which each client computes gradients locally using a batch size of 128.
The server averages the uploaded gradients for global model updates with a learning rate of 0.01.
Input data is normalized within the 0 to 1 range.

\textbf{Attack setting.}
\revise{According to \cite{grad_inversion_survey}, a batch size of 1 in the first few training rounds leads to the most effective attack scenario, as smaller batch sizes can reduce the search space and increase attack effectiveness.
Motivated by this finding, we set the batch size to 1 to assess defense performance under worst-case scenarios.
Here, clients craft uploaded gradients for each sample with a certain defense.
The server then exploits the uploaded gradients to reconstruct clients' raw data and we report the average quality of 1000 reconstructed data.
See Figure \ref{fig_round_impact} (Appendix \ref{appendix_round_impact}) for the defense performance of \sysname over different rounds.}

\section{Empirical Evaluation}
\label{exp}

\begin{figure}[!t]
	\centering
		\includegraphics[width=0.95\linewidth]{images/attack_res/legend.png}
        \\
	\subfigure[SVHN]{
		\includegraphics[width=0.48\linewidth]{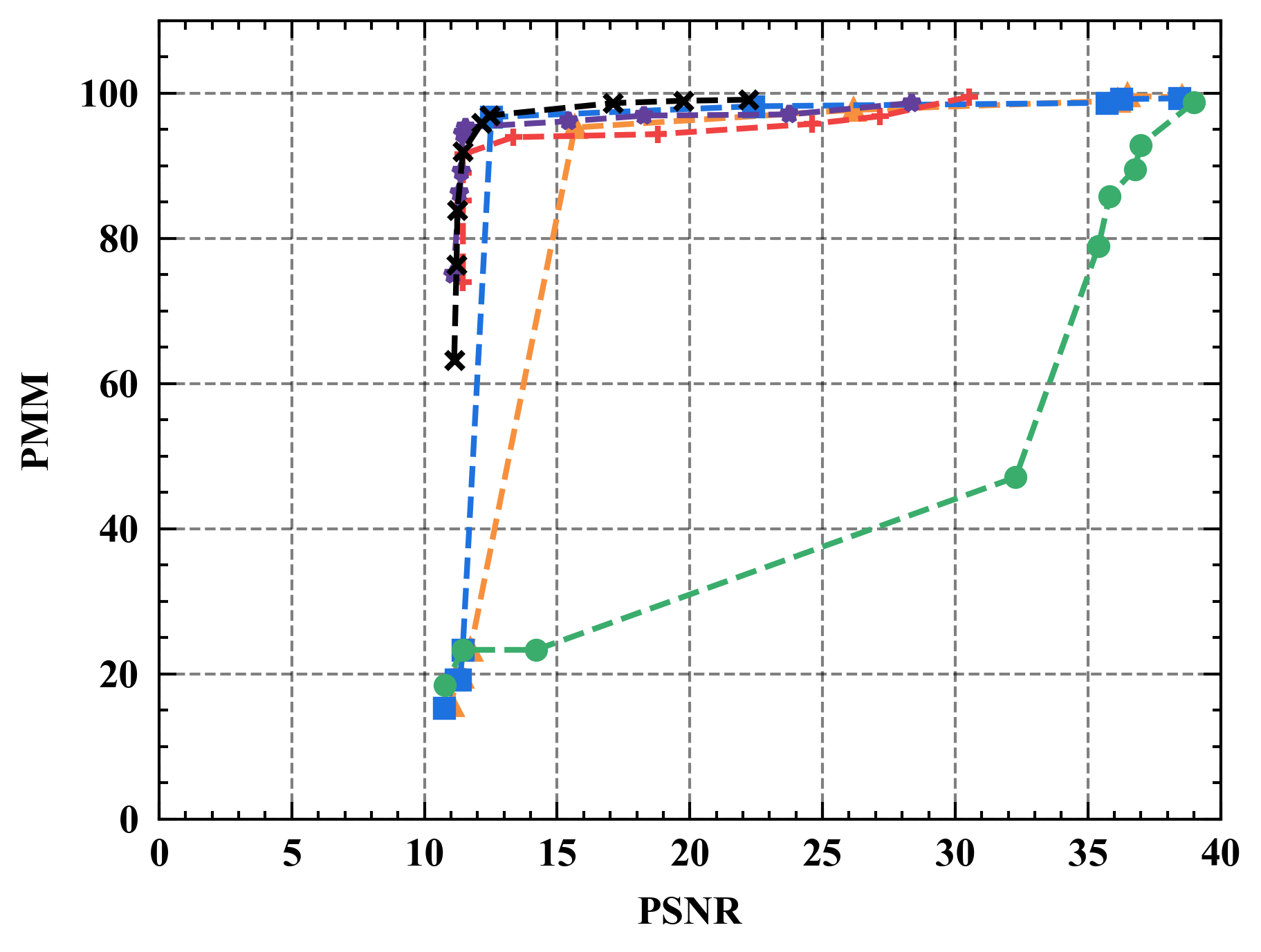}}
	\subfigure[CIFAR100]{
		\includegraphics[width=0.48\linewidth]{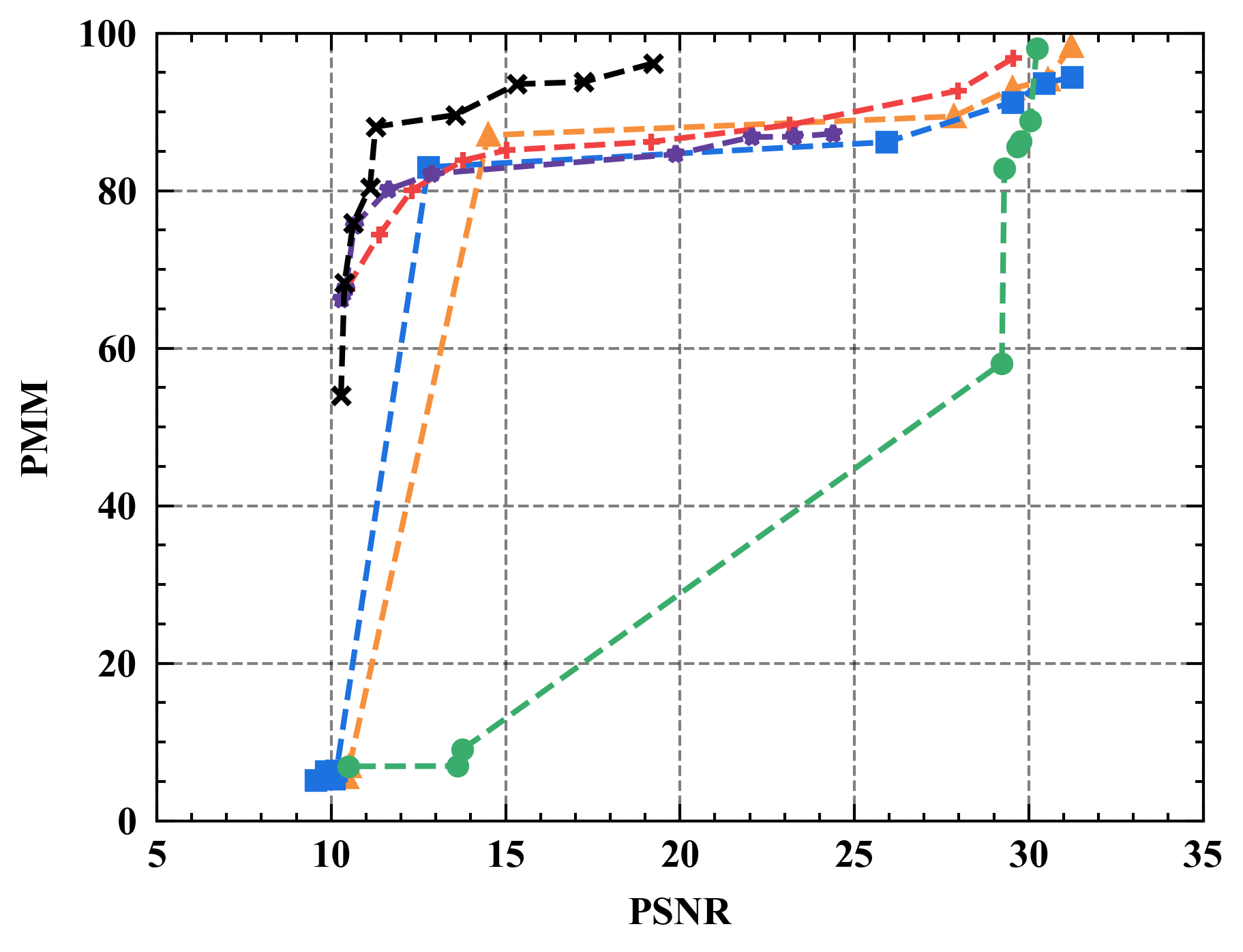}}
	\caption{The performance of various defenses against InvertingGrad in LetNet trained with SVHN and CIFAR100.}
	\label{exp_diff_dataset} 
\end{figure}

\begin{figure}[!t]
	\centering
		\includegraphics[width=0.95\linewidth]{images/attack_res/legend.png}
        \\
	\subfigure[InvertingGrad]{
		\includegraphics[width=0.48\linewidth]{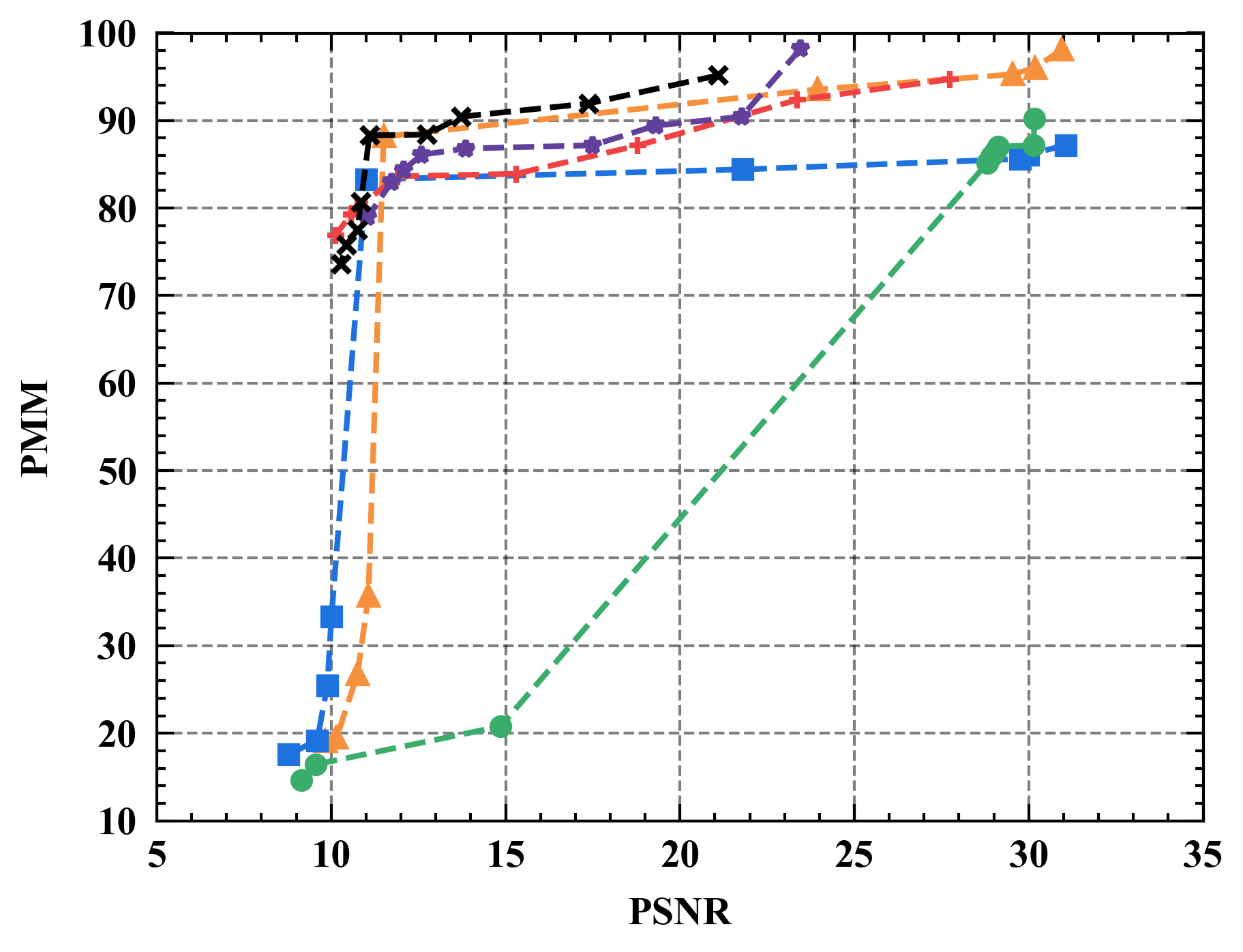}}
	\subfigure[GGL]{
		\includegraphics[width=0.48\linewidth]{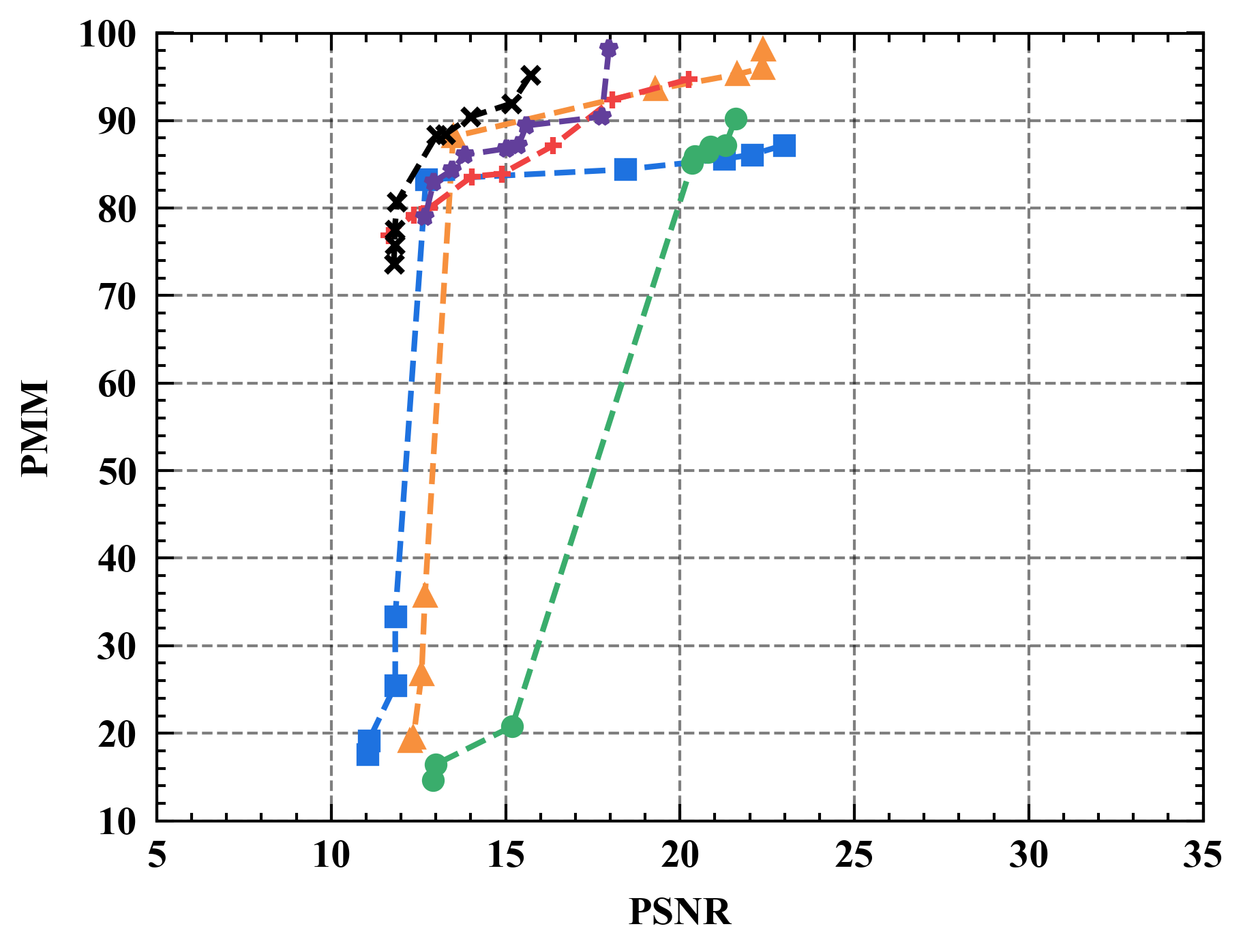}}
	\caption{The performance (PMM-PSNR curves) of defenses against InvertingGrad in LetNet with Non-IID setting.}
	\label{exp_non_iid} 
\end{figure}

\subsection{Privacy-Utility Trade-off}

\textbf{Numerical results.}
Figure \ref{res_lenet_cifar10_psnr} depicts the trade-off curves between PMM (utility) and PSNR (privacy) for various defenses against four state-of-the-art attacks on CIFAR10.
Overall, \sysname consistently outperforms baselines by achieving lower PSNR while maintaining comparable PMM.
Moreover, we observe that the impact on model performance by \sysname, Soteria, and Pruning is relatively mild compared to other defenses.
Particularly, excessively aggressive defense strengths, such as a noise magnitude of $10^{-2}$ or above for DP and a quantization precision of 8 bits or fewer for GQ, can significantly deteriorate the model's performance.

\textbf{Semantic results.}
We observe that the reconstruction quality of GGL, as measured by PSNR, is significantly inferior to the other three attacks.
One possible explanation is that GGL focuses more on semantics rather than exact pixel-level detail.
To substantiate this conjecture, Figure \ref{recovered_image_visualization} showcases the reconstruction images of GGL.
Wherein, GGL indeed recovers the privacy information embedded in the original image (i.e., airplane), highlighting its attack effectiveness.
Furthermore, compared to the PSNR metric, LPIPS places greater emphasis on the similarity of DNN-extracted features, thereby underscoring the importance of semantic-level similarity over pixel-level similarity.
Therefore, we deploy LPIPS metric and the evaluation network for further assessment of GGL's effectiveness.
From Figure \ref{exp_GGA}(a), we observe that the attack performance of GGL is competitive with InvertingGrad against Soteria and \sysname.
Intriguingly, in Figure \ref{exp_GGA}(b), the evaluation network consistently rates GGL's attack effectiveness highly across varying defense strengths.
This aligns with our expectations, given that GGL's output images are noise-free, as displayed in Figure \ref{recovered_image_visualization}.
The above analysis stresses the necessity of employing diverse privacy metrics. 

\textbf{Multi-Metric evaluation.}
Figure \ref{exp_diff_metric} shows the performance of defenses under various privacy metrics.
As can be seen, \sysname achieves a superior trade-off between privacy and utility across a multiplicity of evaluation metrics.
For instance, \sysname enables the inclusion of approximately 30\% noise information in constructed data with negligible impact on the model's performance, whereas Soteria achieves only around 20\% noise inclusion capability.

\textbf{Safeguarding different model architectures.}
Figure \ref{exp_diff_model} presents the performance of various defenses when applied to ResNet and ViT.
Remarkably, \sysname exhibits a superior trade-off between utility and privacy, particularly manifest when implemented on larger-scale models.
We hypothesize that this can likely be attributed to the increased non-linearity capacity of large models, rendering the underlying assumptions of other defenses less valid and consequently leading to degraded effectiveness.

\textbf{Safeguarding different datasets.}
We extend our examination to alternative datasets, namely SVHN and CIFAR100.
The results, as shown in Figure \ref{exp_diff_dataset}, not only confirm the consistent superiority of \sysname' performance, but also reveal two noteworthy observations.
Firstly, we observe a better reconstruction quality for attacks carried out on SVHN, with the highest achieved PSNR approaching 40.
This is probably because SVHN is a less sophisticated dataset, rendering the construction easier.
Secondly, we observe the remarkable effectiveness of \sysname compared to baselines when applied to the CIFAR100 dataset, suggesting that \sysname exhibits greater resilience in handling more complex datasets.

\textbf{Defense with Non-IID setting.}
In the Non-IID setting, the datasets of clients are heterogeneous, posing greater challenges to maintaining performance.
We use client-label-imbalance configuration and employ Dirichlet distribution with a concentration parameter of 1 to allocate fractions of each class label to different clients.
We then perform without-replacement sampling from the corresponding original training datasets for each client, guaranteeing alignment with this allocation.
Our empirical results, shown in Figure \ref{exp_non_iid}, shows that \sysname maintains its edge over the baselines even under the Non-IID setting. 
Moreover, while we observe little change in the quality of attacker-reconstructed images under non-IID conditions, the difficulty of maintaining performance increases significantly compared to the IID setting.

\begin{table}[t]
\centering
\caption{The elapsed time in seconds (s) for a client per round with varying defenses using an A10 GPU. We use a batch size of 128. "Ours + Warm-Start" refers to \sysname with the warm-start strategy, where robust data from previous rounds is reused to reduce the iteration count (halving $\iota$).}
\label{time_comp_exp}
\scriptsize
\begin{tabular}{@{}cccccc@{}}
\toprule
Dfense & LeNet & ResNet10 & ResNet18 & ResNet34 & ViT \\ \midrule
Pruning & 0.0187 & 0.0495 & 0.0873 & 0.1370 & 1.1232 \\
Gaussian & 0.0467 & 0.7317 & 1.8408 & 3.3020 & 15.5051 \\
Laplace & 0.0469 & 0.7322 & 1.8414 & 3.3025 & 15.5236 \\
GQ & 0.0179 & 0.0277 & 0.0514 & 0.1050 & 0.4489 \\
Soteria & 0.4103 & 11.3148 & 21.7539 & 39.4410 & 347.5668 \\
Ours & 0.3340 & 1.7341 & 2.8539 & 4.7012 & 30.2268 \\ 
Ours + Warm-Start & 0.1418 & 0.8523 & 1.4262 & 2.3548 & 15.0234 \\ \bottomrule
\end{tabular}
\end{table}

\begin{table}[t]
\centering
\caption{The PSNR and PMM achieved by \sysname over varying $\beta$ against InvertingGrad in LeNet trained with CIFAR10.}
\label{impact_beta}
\scriptsize
\begin{tabular}{@{}cccccccc@{}}
\toprule
$\beta$ & $10^{-3}$ & $10^{-2}$ & $10^{-1}$ & $10^{0}$ & $10^{1}$ & $10^{2}$ & $10^{3}$ \\ \midrule
PSNR & 13.89 & 13.62 & 13.44 & 13.20 & 13.08 & 12.70 & 12.04 \\
PMM & 91.93 & 91.12 & 91.06 & 90.69 & 90.00 & 89.30 & 88.90 \\ \bottomrule
\end{tabular}
\end{table}

\begin{table}[t]
\centering
\caption{The PSNR and PMM achieved by \sysname over varying $\iota$ against InvertingGrad in LeNet trained with CIFAR10.}
\label{impact_i}
\scriptsize
\begin{tabular}{@{}ccccc@{}}
\toprule
$\iota$ & 5 & 10 & 15 & 20 \\ \midrule
PSNR & 13.05 & 13.20 & 13.37 & 13.45 \\
PMM & 85.61 & 90.69 & 91.19 & 92.17 \\ \bottomrule
\end{tabular}
\end{table}

\subsection{Visual Analysis}
\label{exp_visual}

\begin{figure}[!t]
\centering
\includegraphics[width=0.65\linewidth]{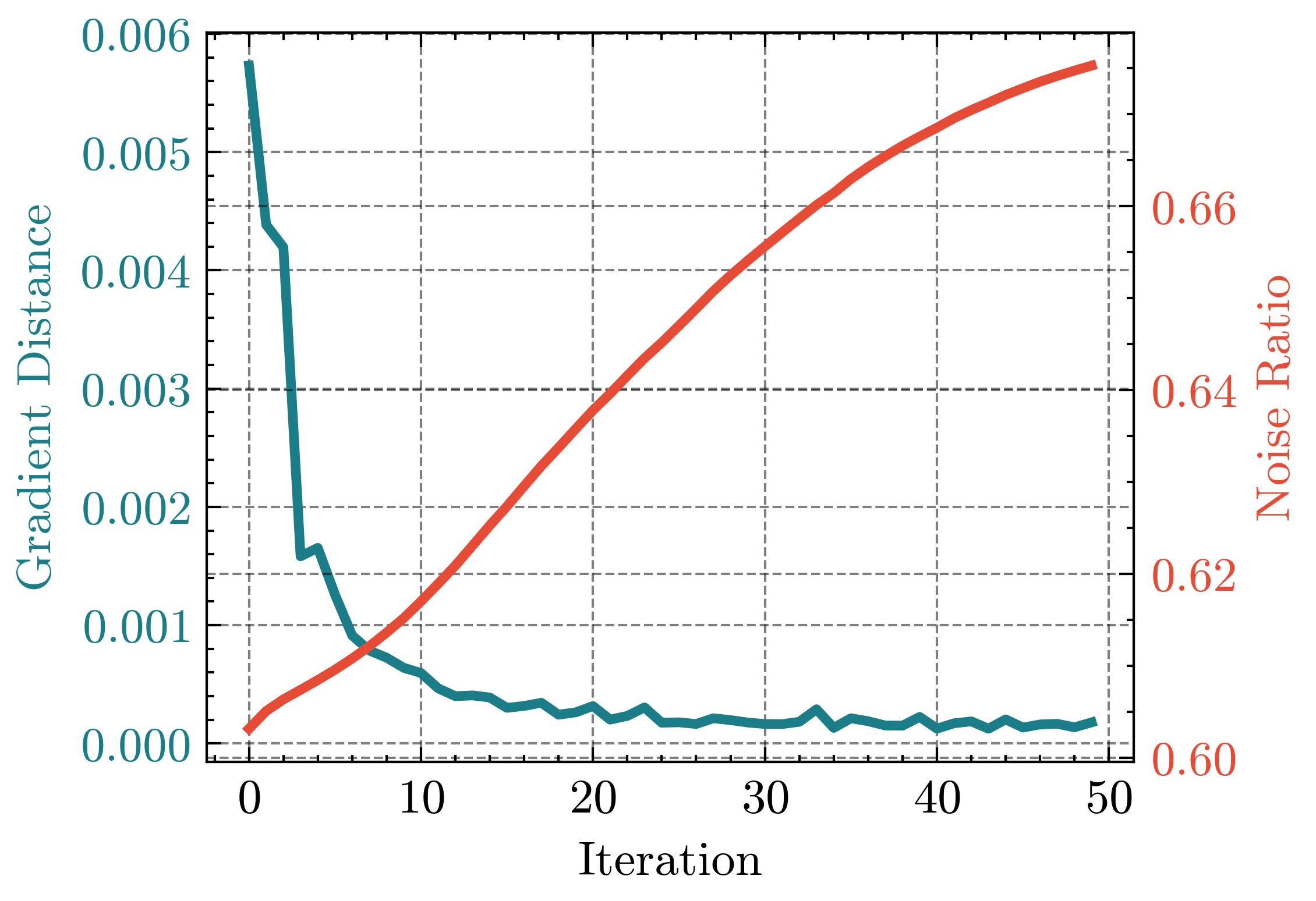}
\caption{\revise{Gradient distance and noise ratio trends across iterations.}}
\label{fig:grad_trend}
\end{figure}

\begin{figure}[!t]
	\centering
	\subfigure[Raw Data and Robust Data]{
		\includegraphics[width=0.4\linewidth]{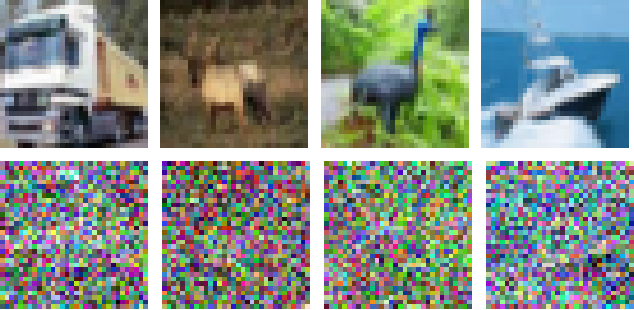}}
	\subfigure[Recovered Images]{
		\includegraphics[width=0.4\linewidth]{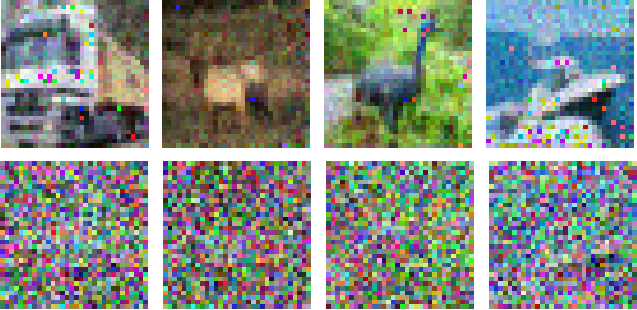}}
	\caption{\revise{(a) original data (top row) and robust data (bottom row). (b) gradient-reconstructed counterparts.}}
	\label{fig:vis_samples}
\end{figure}

\revise{To intuitively understand \sysname's effectiveness, we provide qualitative analyses here.
Figure \ref{fig:grad_trend} shows the gradient distance between robust and original data alongside the evaluation network's predicted noise ratio across iterations.
As shown, \sysname achieves convergence within 10 iterations, with gradient distances stabilizing after 20 iterations.
Figure \ref{fig:vis_samples} visualizes four randomly selected CIFAR-10 samples: (a) original vs. robust data (10 iterations), and (b) images reconstructed from gradients of raw data and robust data using InvertingGrad.
The robust data appears as pure noise to the human eye.
According to our theoretical analysis, the corresponding reconstructed data should also exhibit noise characteristics, which Figure \ref{fig:vis_samples}(b) verifies.
Furthermore, we observe that reconstructed images exhibit strong correlations with their robust counterparts.
In detail, the third sample shows green tones, and the last sample preserves blue hues.}

\begin{figure}
    \centering
    \includegraphics[width=0.6\linewidth]{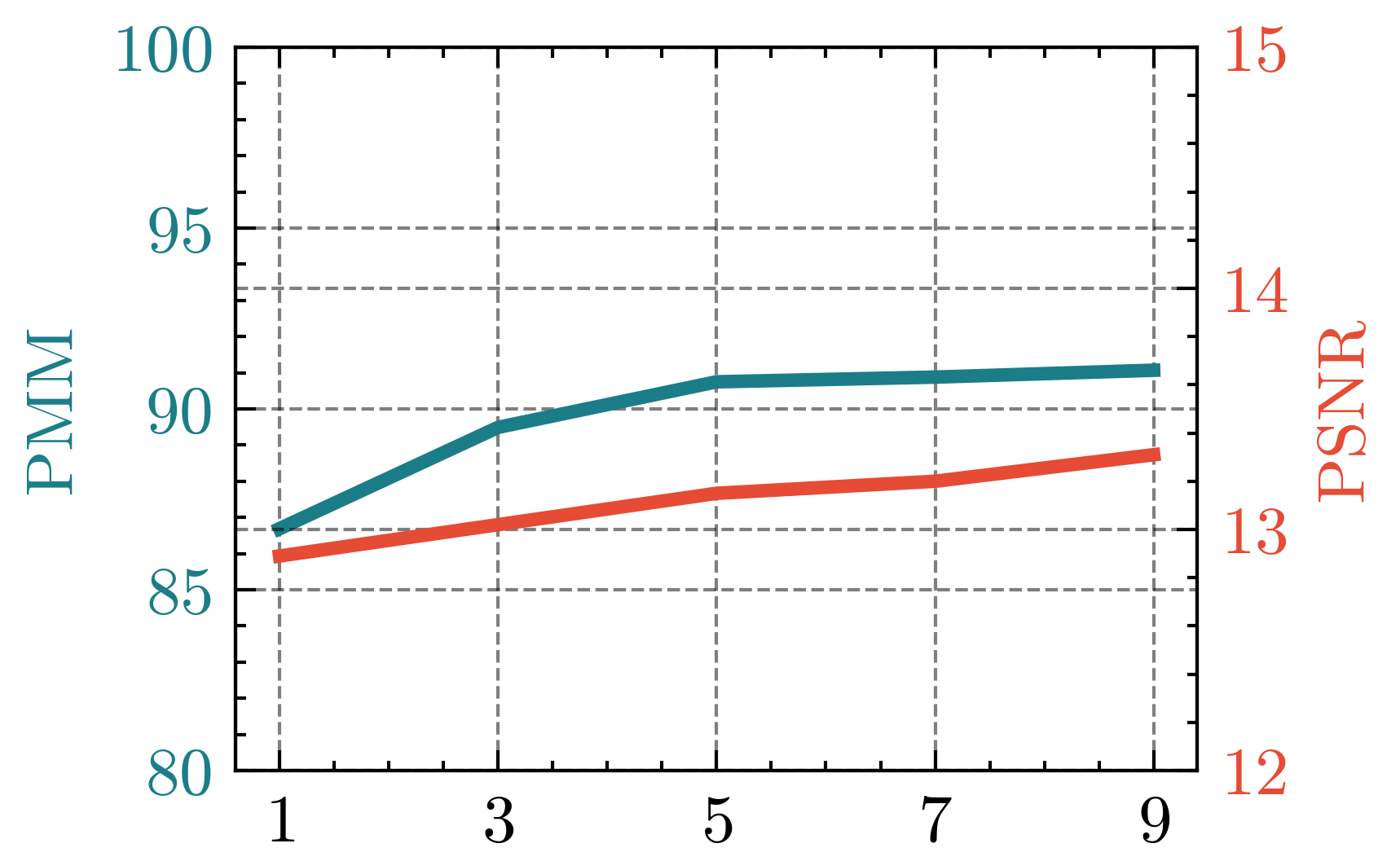}
    \caption{\revise{Impact of warm-start strategy on \sysname's performance under varying $\iota$ values. We InvertingGrad on LeNet trained with CIFAR-10.}}
    \label{fig_warmup_impact}
\end{figure}

\subsection{Empirical Time Complexity}
\label{exp_time_complex_comp}

The additional time cost imposed by defenses is a crucial dimension in evaluating their practicality.
Table \ref{time_comp_exp} compares the runtime required for a single client per round when employing different defenses.
While \sysname and Soteria incur similar time costs for compact models like LeNet, both are an order of magnitude higher than lightweight methods like Pruning, Gaussian, Laplace, and Gradient GQ.
This is due to the fact that \sysname and Soteria require multiple forward-backward passes to search for better utility-privacy trade-offs.

When scaling up to larger models including ResNet and ViT, the execution time for Pruning and GQ scales almost linearly.
\sysname, Gaussian, and Laplace present time expenses of the same magnitude.
Notably, Soteria makes a significantly higher time cost of 40s on ResNet34, which is 8.4$\times$ higher than the time required for \sysname.
The heavy overhead of Soteria stems from the intensive execution of forward-backward passes per neuron within the fully connected layers.
As a result, as the number of fully connected neurons increases in large models, the time cost of Soteria escalates drastically.
In contrast, \sysname’s time cost depends on the predefined number of iterations.
Empirical results indicate that \sysname achieves satisfactory performance within merely 10 iterations, far lower than the overhead associated with Soteria.
\revise{Moreover, we introduce a warm-start strategy to further reduce \sysname’s overhead, which reuses robust data generated in earlier rounds as initialization for subsequent iterations (excluding the first round for generating robust data for $x$).
Figure \ref{fig_warmup_impact} evaluates the impact of this strategy on \sysname’s performance across different $\iota$ values.
We find that reducing $\iota$ to 5 just results in tiny performance degradation: PMM decreases slightly from 90.69 to 90.45, while PSNR increases marginally from 13.20 to 13.27 on LeNet trained with CIFAR-10 against InvertingGrad.
Therefore, we set $\iota$ to 5 after employing the warm-start strategy to reassess the time cost, as shown in Table \ref{time_comp_exp}.
This makes \sysname significantly more efficient than Soteria, while maintaining a better utility-privacy trade-off than DP, GQ, and gradient pruning.
Additionally, we estimate \sysname’s overhead on edge devices to be within acceptable limits (Appendix \ref{appendix_overhead_edge_device} provides details).}

\begin{table}[t]
\centering
\caption{The PSNR and PMM achieved by \sysname over varying $\alpha$ against InvertingGrad in LeNet trained with CIFAR10.}
\label{impact_alpha}
\scriptsize
\begin{tabular}{@{}ccccccc@{}}
\toprule
$\alpha$ & 0 & 0.1 & 0.3 & 0.5 & 0.7 & 0.9 \\ \midrule
PSNR & 21.24 & 16.52 & 14.99 & 13.20 & 12.53 & 10.52 \\
PMM & 98.52 & 95.45 & 92.61 & 90.69 & 90.08 & 89.34 \\ \bottomrule
\end{tabular}
\end{table}

\begin{table}[t]
\centering
\caption{The PSNR and PMM achieved by \sysname over varying $\tau$ against InvertingGrad in LeNet trained with CIFAR10.}
\label{impact_tau}
\scriptsize
\begin{tabular}{@{}cccccc@{}}
\toprule
$\tau$ & 0.91 & 0.93 & 0.95 & 0.97 & 1.00 \\ \midrule
PSNR & 12.93 & 13.05 & 13.20 & 13.25 & 15.44 \\
PMM & 90.98 & 90.82 & 90.69 & 90.46 & 89.28 \\ \bottomrule
\end{tabular}
\end{table}

\subsection{Comparison to Cryptographic Methods}

\begin{table}[!t]
\caption{\revise{Performance of cryptographic methods \cite{secure_agg} under InvertingGrad attacks. PSNR values correspond to batch sizes of 1 and 128.}}
\label{crypto_comp}
\scriptsize
\centering
\begin{tabular}{@{}cccc@{}}
\toprule
Model    & PMM & PSNR & Time (Client/Server)                  \\ \midrule
LeNet    &  99.91 &  18.67/10.24    & 2.0784s/2.3185s       \\
ResNet10 & 100.24 &  15.58/10.39    & 642.6589s/719.8361s   \\
ResNet18 & 100.05 &  15.52/10.08    & 1460.1683s/1635.4287s \\ \bottomrule
\end{tabular}
\end{table}

\revise{Table \ref{crypto_comp} compares the performance of secret-sharing-based cryptographic methods \cite{secure_agg} against InvertingGrad in CIFAR-10, reporting PSNR values for batch sizes of 1 and 128.
Cryptographic methods impose computational overhead not only on clients but also on the server, so we report the total time per round for both client and server.
Compared to perturbation-based methods (e.g., \sysname in Table \ref{time_comp_exp}), cryptographic methods incur much higher costs.
For LeNet, cryptographic methods are 6$\times$ slower than \sysname without warm-start and 14.6$\times$ slower with warm-start.
On larger models like ResNet, the overhead becomes orders of magnitude greater, reaching hundreds of times the cost of \sysname.
While cryptographic methods maintain near-perfect PMM (e.g., 99.91–100.24), their prohibitive time costs severely limit practicality.
Notice that homomorphic encryption-based methods are not included due to two key factors \cite{WenZLCCZ23}.
First, they incur significantly higher computational costs than secret-sharing-based methods.
Second, homomorphic encryption can enable the server to perform algebraic operations directly on encrypted parameters without decryption \cite{park2022privacy}, which inherently eliminates the threat model assumed by GLAs.}

\subsection{Sub-components Analysis}

In this subsection, we investigate the impact of four sub-components of \sysname on its performance.
We fix $\epsilon=0.1$.

\textbf{Impact of balance factor $\beta$.}
Table \ref{impact_beta} reports the performance of \sysname as $\beta$ varies against InvertingGrad.
Overall, the higher $\beta$ is, the lower PMM is, and the higher MSE is.
Moreover, the performance of \sysname is more robust to adjustments in $\beta$ as compared to $\alpha$, with negligible performance fluctuations across different values of $\beta$.
Thus, maintaining $\beta$ at 1 may be a practical default.

\textbf{Impact of iterations $\iota$.}
Table \ref{impact_alpha} reports the performance of \sysname when configured with varying $\alpha$ against InvertingGrad.
Increasing iterations from 5 to 10 significantly bolsters \sysname's performance, enabling the acquisition of higher PMM with similar PSNR.
Intuitively, early iterations tend to yield higher marginal benefits as the algorithm rapidly approaches convergence.
Beyond a certain number of iterations (here identified as 10 iterations) is surpassed, \sysname reaches a point of saturation, beyond which significant returns are no longer generated.
Hence, from both a performance and computational perspective, an approximation of 10 iterations serves as a practical choice.

\textbf{Impact of $\alpha$.}
A higher value of $\alpha$ suggests that the initialized images drift further from the clients' data, enhancing privacy at the potential cost of reduced utility.
The results in Table \ref{impact_alpha} align with this thought where a rise in $\alpha$ corresponds to lower PSNR and PMM values.
As analyzed in Section \ref{solving_opt_problem}, when $\alpha$ is set to 0 (without noise-blended initialization), the PSNR surges, indicating high semantic similarity between robust and clients' data.
To mitigate adversarial vulnerability, we employ noise-blended initialization, enabling a significant decrease in PSNR with a moderate sacrifice in PMM.
For example, by adjusting $\alpha$ from 0 to 0.1, we achieve a substantial reduction in PSNR by approximately 5, while the decline in PMM remains modest, estimated at only around 3.

\textbf{Impact of $\tau$.}
$\tau$ shapes the alignment between the gradients of client data and robust data.
Table \ref{impact_tau} reports the results of tuning $\tau$.
By including $\tau$ into \sysname (transitioning from 1 to 0.97), a marginal PMM reduction of 1\% is observed while simultaneously securing a considerable 2-point decrease in PSNR.
Further increases in $\tau$ yield negligible benefits, suggesting that the observed improvements reach a plateau.

\revise{\textbf{Impact of element-wise weight.}
Table \ref{impact_element_weight} compares the performance of \sysname with and without element-wise weight.
While the element-wise weight introduces only a small increase in PSNR (from 13.15 to 13.20), it achieves a significant improvement in PMM (from 85.32 to 90.69), demonstrating its critical role in \sysname.}

\begin{table}[t]
\centering
\caption{\revise{The performance of \sysname with and without element-wise weight against InvertingGrad in LeNet trained with CIFAR10.}}
\label{impact_element_weight}
\scriptsize
\begin{tabular}{@{}ccc@{}}
\toprule
\sysname & PSNR & PMM \\ \midrule
w.o. Element-wise Weight & 13.15 & 85.32 \\
w. Element-wise Weight & 13.20 & 90.69  \\ \bottomrule
\end{tabular}
\end{table}

\begin{figure}[!t]
\centering
\subfigure[w.o. Evaluation Network]{
\includegraphics[width=0.48\linewidth]{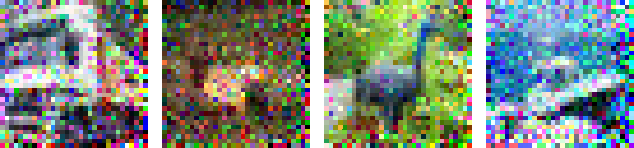}}
\subfigure[Reconstructions]{
\includegraphics[width=0.48\linewidth]{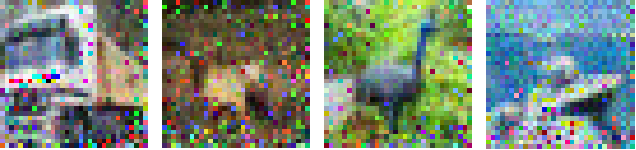}}
\caption{\revise{Visualization of robust data generation and InvertingGrad reconstructions when replacing the evaluation network with MSE distance metric.}}
\label{fig:evaluation_network_impact}
\end{figure}

\revise{\textbf{Impact of the evaluation network.}
Figure \ref{fig:evaluation_network_impact} visualizes robust data generated by \sysname and InvertingGrad reconstructions when replacing the evaluation network with MSE distance metric.
Compared to Figure \ref{fig:vis_samples}, the MSE-based \sysname focuses on altering specific pixels rather than disrupting global semantic structures.
This results in reconstructed images retaining most semantic information from the original data.
The gradient distance between the generated robust data and real data remains approximately 0.001, achieving a PMM score of 91.05.}

\section{Discussion, Conclusion, and Future Work}
\label{discussion}
\textbf{Why does \sysname work?}
We are interested in why \sysname is more effective than perturbation-based methods.
In fact, we discover that perturbation-based methods are theoretically optimal under certain assumptions, which, unfortunately, do not often hold for DNNs.
To explicate, perturbing each gradient element equally (DP and GQ) assumes that every gradient element carries equal privacy information and contributes equally to the model performance, making uniform perturbation theoretically optimal under this assumption.
By considering all gradient elements as possessing identical privacy information, the optimal solution derived from Taylor expansion is the preservation of gradients with large magnitudes, \ie gradient pruning.
Soteria improves on gradient pruning by estimating the privacy information of each gradient element based on the \textit{linearity assumption} of DNNs, selectively removing those with the highest privacy information.

\sysname uploads gradients generated by robust data.
By reducing semantic similarity between the robust data and clients' data, the uploaded gradients contain less privacy information about clients' data.
Importantly, such privacy estimation fashion avoids potential estimation errors that may arise from questionable assumptions made by perturbation-based methods.
Regarding utility, \sysname prioritizes important parameters and narrows the gap between the gradients of these parameters with respect to robust and clients' data.
Section \ref{privacy_metric} validates that preserving the gradient of important parameters is a better solution than retaining gradients with the largest magnitude.
These insights explain \sysname's superior performance over perturbation-based defenses.

\textbf{Conclusion and future work.}
\revise{Overall, we believe that \sysname brings a novel idea for privacy protection by shifting the focus from gradient perturbation or cryptographic approaches to data-centric defenses.
Our empirical validation also demonstrated that \sysname achieves superior privacy-utility tradeoffs compared to existing methods.}
The main drawback of \sysname is its higher computational demand.
However, this additional cost is justified by the enhanced performance and is manageable in typical resource-constrained settings like IoT (Appendix \ref{appendix_overhead_edge_device}).
Looking ahead, we aim to reduce the computational burden.

\bibliographystyle{plainnat}
\bibliography{references}

\appendix

\section{Weight Comparison}
\label{weighting_mechanism_comparison}

Common metrics for parameter importance include Neuron Shapley \cite{neuron_shapley} and Integrated Gradient \cite{integrated_grad}.
While Neuron Shapley theoretically captures neuron contributions to model performance, its prohibitive computational cost limits practicality \cite{neuron_shapley}.
Here, parameters of neurons considered insignificant by Neuron Shapley are deemed non-important.
We train a LeNet model on the CIFAR10 dataset for 20 epochs (20 epochs, learning rate 0.01, batch size \revise{128}).
We utilize these two metrics and our metric to differentiate important parameters and subsequently eliminate gradients (80\% pruning rate) from neurons deemed non-important.
Results in Table \ref{metric_comp} show that our metric achieves competitive performance while requiring significantly less computational overhead compared to Neuron Shapley and Integrated Gradient.

\section{Validation of SSIM and LPIPS}
\label{validation_ssim_lpips}

\begin{table}[!t]
\caption{The performance and overhead of different metrics.}
\label{metric_comp}
\centering
\scriptsize
\begin{tabular}{@{}c|ccc@{}}
\toprule
 Metric & Accuracy (\%) & Time (s) \\ \midrule
Integrated Gradient & 49.54 & 2155.85 \\
Shapley Value & 49.50 & 193672.67 \\
Our Metric & 49.55 & 209.20 \\
\bottomrule
\end{tabular}
\end{table}

\begin{figure}[!t]
\centering
\subfigure[SSIM]{
\includegraphics[width=0.48\linewidth]{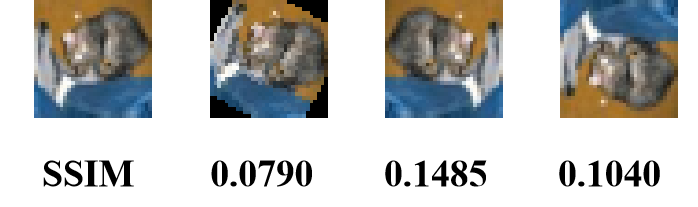}}
\subfigure[LPIPS]{
\includegraphics[width=0.48\linewidth]{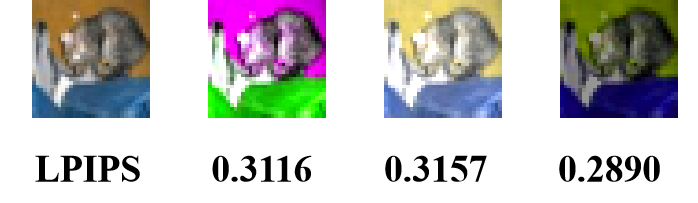}}
\caption{(a) Rotation and flipping. (b) Color jitter.}
\label{ssim_demon}
\end{figure}

We here demonstrate that SSIM and LPIPS fail to align with semantic similarity, rendering them unsuitable for privacy evaluation.
Figures \ref{ssim_demon}(a) and \ref{ssim_demon}(b) apply semantic-preserving transformations (e.g., rotation, flipping, color jitter) to original images, resulting in drastic metric changes: SSIM drops to 0.1 with rotation, and LPIPS increases to 0.3 with color jitter.
While these values suggest extreme dissimilarity, the transformed images retain semantic equivalence to their originals.
This highlights that SSIM and LPIPS are inappropriate for privacy metrics.

\section{Proof of Privacy Metric}
\label{proof_of_privacy_metric}

A convex function $g(x)$ can be represented using its conjugate form as follows:
\begin{equation}
\label{proof_privacy_equ_1}
\begin{split}
    &g(x) = \Max_{t} \ tx - g(t), \\
    &t = g'(x), g(t) = -f(x) + f'(x) x,
\end{split}
\end{equation}
where $t$ belongs to the range of $g'(x)$.
Equation \ref{proof_privacy_equ_1} indicates that the value of $g(x)$ at a certain point can be evaluated by maximizing $t$.
When given an $x$, there exists a fixed corresponding value for $t$, which we denote as $t=H(x)$.
Consequently, Equation \ref{proof_privacy_equ_1} can be rewritten as follows:
\begin{equation}
\label{proof_privacy_equ_2}
\begin{split}
    g(x) = \Max_{H(x)} \ H(x) x - g(H(x)),
\end{split}
\end{equation}
where the range of $H(x)$ belongs to the range of $g'(x)$.
The JS divergence of $p(x)$ and $q(x)$ can be written as
\begin{equation}
\label{proof_privacy_equ_3}
\begin{split}
\nonumber
    &\frac{1}{2} \int p(x) log \frac{2p(x)}{p(x)+q(x)} + q(x) log \frac{2q(x)}{p(x)+q(x)} dx \\
    &= \frac{1}{2} \int p(x) \{ log \frac{2}{1+q(x)/p(x)} + 
    \frac{q(x)}{p(x)} log \frac{2q(x)/p(x)}{1+q(x)/p(x)}
    \}
    dx.
\end{split}
\end{equation}
Since $h(z)$ is a convex function, we can represent JS divergence using its conjugate form as follows:
\begin{equation}
\label{proof_privacy_equ_4}
\begin{split}
    &\frac{1}{2} \int p(x) log \frac{2p(x)}{p(x)+q(x)} + q(x) log \frac{2q(x)}{p(x)+q(x)} dx \\
    &= \frac{1}{2} \int p(x) h(\frac{q(x)}{p(x)}) dx \\
    &= \frac{1}{2} \int p(x) \{ \Max_{H(\frac{q(x)}{p(x)})} \ H(\frac{q(x)}{p(x)}) \frac{q(x)}{p(x)} - g(H(\frac{q(x)}{p(x)})) \} dx \\
    &= \frac{1}{2} \Max_{H(\frac{q(x)}{p(x)})} \int q(x) H(\frac{q(x)}{p(x)}) - p(x) g(H(\frac{q(x)}{p(x)})) dx.
\end{split}
\end{equation}
According to Equation \ref{proof_privacy_equ_2}, the corresponding form of $g(\cdot)$ to $h(\cdot)$ is $g(t)=-log(2-e^t)$.
Substituting the specific form of $g(t)=-log(2-e^t)$ into Equation \ref{proof_privacy_equ_4}, we obtain:
\begin{equation}
\label{proof_privacy_equ_5}
\begin{split}
\nonumber
    &\frac{1}{2} \Max_{H(\frac{q(x)}{p(x)})} \int q(x) H(\frac{q(x)}{p(x)}) - p(x) g(H(\frac{q(x)}{p(x)})) dx\\
    & = \frac{1}{2} \Max_{H(\frac{q(x)}{p(x)})} \int q(x) H(\frac{q(x)}{p(x)}) + p(x) log(2-e^{H(\frac{q(x)}{p(x)})}) dx.
\end{split}
\end{equation}
Setting $H(\frac{q(x)}{p(x)})=log(2D(x))$ and substituting it into the ebove equation, we obtain:
\begin{equation}
\label{proof_privacy_equ_6}
\begin{split}
\nonumber
    \frac{1}{2} \Max_{D(x)} \mathbb{E}_{x \sim p(x)} [log (1-D(x))] +  \mathbb{E}_{x \sim q(x)} [log D(x)]
    + log2.
\end{split}
\end{equation}
Moreover, the definition of conjugate function requires $H(\cdot)$ to be limited to the range of $g'(t)$.
Since the range of $g'(t)$ is smaller than $log2$, we can derive that the range of $D(\cdot)$ is constrained to the interval of 0 to 1.

\begin{table}[!t]
\centering
\caption{Total variance (TV) quantifies the smoothness of inputs. $L_2$-norm controls the pixel value within a legal range $[0 \sim 1]$. BN statistics force the statistics of recovered images to close the statistics of datasets. KL distance restricts the latent variables to follow Gaussian distribution.}
\label{attack_setting}
\scriptsize
\begin{tabular}{@{}c|cccc@{}}
\toprule
Attack          & iDLG  & InvertingGrad & GradInvertion           & GGL  \\ \midrule
Loss Function & Euclidean & Cosine Similarity & Euclidean & Euclidean \\
Regularization  & None  & TV            & \tabincell{c}{TV+$L_2$-norm\\+BN statistics} & \tabincell{c}{GAN+KL\\ Distance}  \\
Optimizer       & L-BFGS & Adam          & Adam                    & Adam \\
Learning Rate   & 1     & 0.01          & 0.01                    & 0.01 \\
Label Inference & $\checkmark$ & $\checkmark$          & $\checkmark$                    & $\checkmark$ \\
Attack Iteration       & 300   & 4000          & 4000                    & 4000 \\ \bottomrule
\end{tabular}
\end{table}

\section{Detailed Settings}
\label{leaving_detailed_settings}

Table \ref{attack_setting} summarizes the differences between employed attacks and the hyperparameters used.

\textbf{The architecture and training settings of the evaluation network.}
The evaluation network comprises four layers: three convolutional layers (kernel size $3 \times 3$, stride 2, padding 1, with ReLU activation) and a fully connected layer (sigmoid activation).
The number of filters in the convolutional layers is 128, 256, and 512, respectively.
The network predicts noise proportions in inputs and is trained for 20 epochs using Adam optimizer ($10^{-4}$ learning rate, batch size 128).
During training, each batch is mixed with random noise 11 times, with $\alpha \in \{0, 0.1, \dots, 1\}$ before proceeding to the next batch.

\textbf{Non-IID setting.}
We generate non-IID datasets by assigning local data distributions via a symmetric Dirichlet distribution with concentration parameter 1.
For each client, we sample a distribution over data categories, determine the number of samples per label, and randomly select corresponding samples from the training dataset.
This process ensures heterogeneous data distributions across clients, enabling evaluation of defense robustness under non-IID conditions in Section \ref{exp}.

\begin{figure}[!t]
    \centering
    \includegraphics[width=0.5\linewidth]{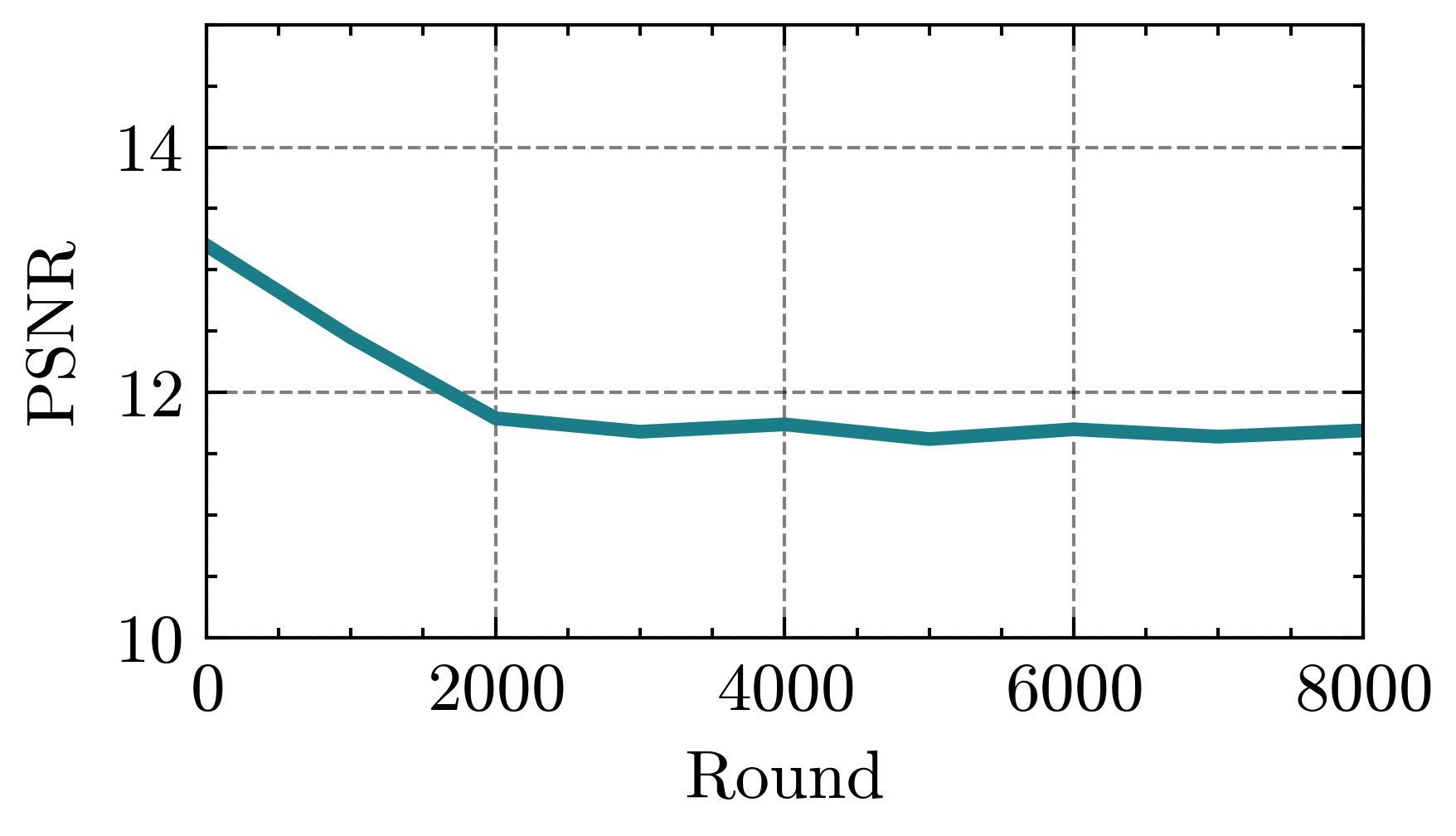}
    \caption{\revise{The performance of \sysname over different rounds.}}
    \label{fig_round_impact}
\end{figure}

\section{\sysname‘s Performance over Rounds}
\label{appendix_round_impact}

Figure \ref{fig_round_impact} presents the performance of \sysname against InvertingGrad across varying rounds.
As the number of rounds increases, \sysname demonstrates better defensive performance.
This is aligned with prior findings that prolonged training reduces model sensitivity to input data, thereby making data reconstruction more challenging.

\section{\sysname on Textual Data}
\label{appendix_applicability_refiner}

\textbf{Overall idea.}
For textual data, meaningless data can be considered as sentences composed of random words.
We train an evaluation network to predict the likelihood that a given piece of text resembles a sentence made up of random words.
Notice NLP models commonly convert discrete tokens into continuous word embeddings.
Since discrete data is challenging to optimize directly, we construct robust data for their word embeddings.
In this scenario, clients no longer upload gradients for the parameters of the first layer, which is less impactful for NLP models since NLP models rarely fine-tune the initial layer.
For data from other modalities, we can define their meaningless data similarly to train evaluation network for \sysname execution.

\textbf{Implementation.}
In practice, we replace a portion of words in a normal sentence with random words, then feed the resulting embeddings into evaluation network.
The network is trained to predict the proportion of random words.
We use DistilBERT fine-tuned on IMDb, fine-tuned on IMDb dataset using Huggingface's default configuration\footnote{\url{https://huggingface.co/lvwerra/distilbert-imdb}}.
The process of generating robust data for word embedding vectors follows the same way used for generating robust data for images.

\textbf{Evaluation.}
We employ BERT and SST-2 dataset using an IID setup with 10 clients and Huggingface's default configuration\footnote{\url{https://huggingface.co/gchhablani/bert-base-cased-finetuned-sst2}}.
Since Soteria does not apply to textual data, we evaluate \sysname's defensive performance against InvertingGrad and TAG~\cite{tag}, comparing it with DP and Pruning.
TAG~\cite{tag} is a GLA specific to textual data, and we follow its original paper's hyperparameters.
We set pruning, DP, and \sysname defenses to strengths of 0.2, 0.001, and $\epsilon = 0.05$, respectively, as these hyperparameters are found to offer the strongest defense while maintaining model performance.
BERT without any defenses achieves an accuracy between 91\% to 92\% on SST-2.
We measure attack performance via precision, recall, and F1 metrics over 1000 reconstructed samples.
For utility, we report the model's accuracy on the test set.
As shown in Table~\ref{eval_on_sst2}, \sysname still achieves better defense performance (lower F1 score) while maintaining model utility.

\begin{table}[!t]
\caption{The performance of three defenses against TAG and InvertingGrad.}
\label{eval_on_sst2}
\centering
\scriptsize
\begin{tabular}{@{}ccccc@{}}
\toprule
Attack                         & Metric         & Pruning & DP-Gaussian & \sysname \\ \midrule
\multirow{2}{*}{TAG}           & Model Accuracy (\%) & 90.7       & 90.6           & 90.9       \\
                               & F1 Score (\%)       & 32.28   & 30.44       & 24.55   \\ \midrule
\multirow{2}{*}{InvertingGrad} & Model Accuracy (\%) & 90.8       & 90.5           & 91.1       \\
                               & F1 Score (\%)       & 50.09   & 49.36       & 35.64   \\ \bottomrule
\end{tabular}
\end{table}

\begin{figure}[!t]
	\centering
	\subfigure[Attribute Inference]{
		\includegraphics[width=0.48\linewidth]{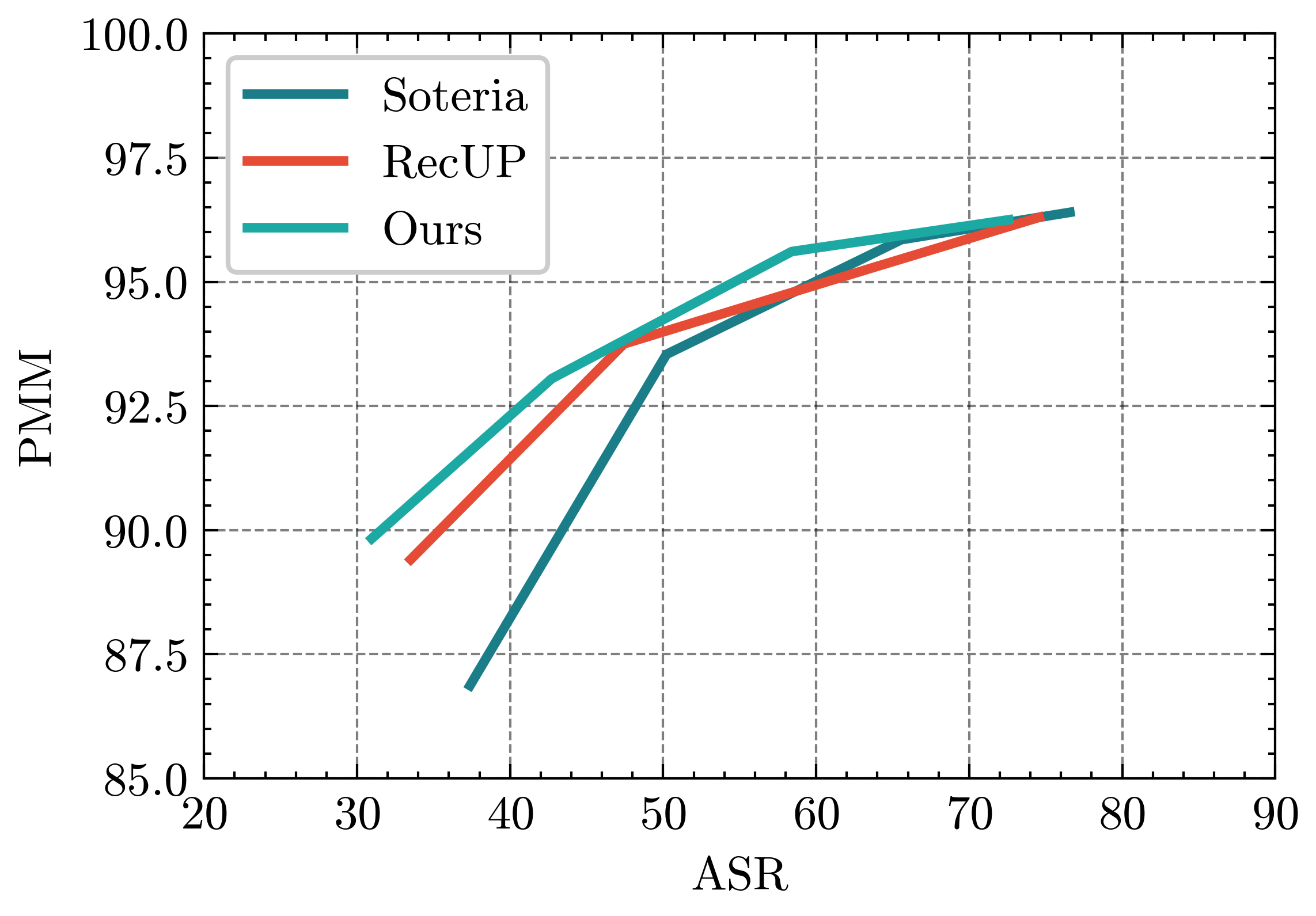}}
	\subfigure[Membership Inference]{
		\includegraphics[width=0.47\linewidth]{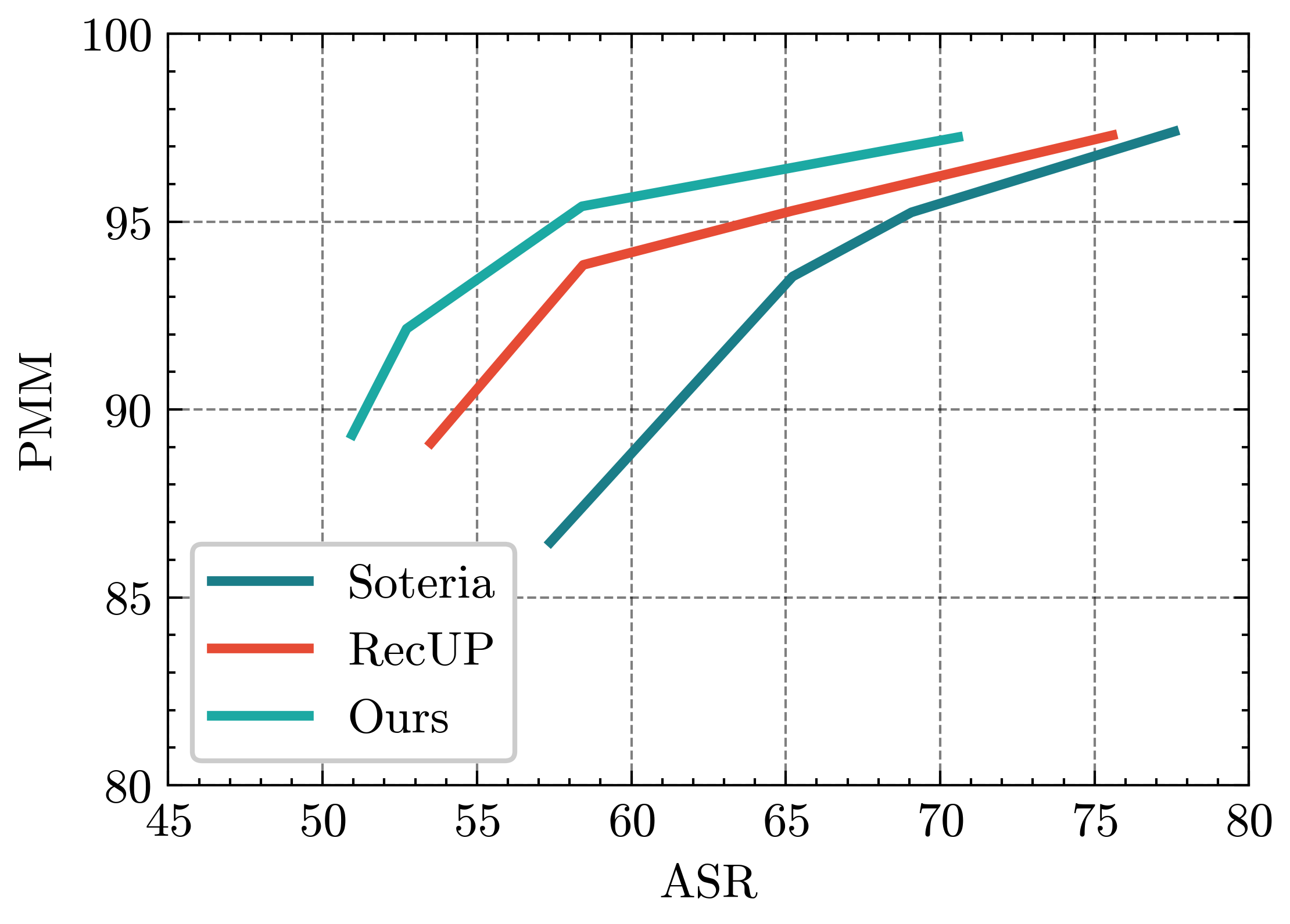}}
	\caption{\revise{The performance of \sysname against attribute and membership inference attacks.}}
	\label{fig_aatr_mia}
\end{figure}

\section{The Effectiveness of \sysname against Attribute and Membership Inference}
\label{appendix_effect_mia_aa}

\revise{Attribute inference attacks try to infer specific attributes from the client's data, while membership inference attacks seek to determine if a specific data point belongs to a client's dataset.
The effectiveness of these attacks primarily arises from models' tendency to overfit their training data rather than from generalizable features.
\sysname promotes semantic divergence between original and robust data during gradient alignment.
This can be intuitively understood as the process of removing task-irrelevant information from the data while preserving features essential for the original task.
From this standpoint, \sysname can effectively mitigate both attribute inference and membership inference attacks.
Following the experimental setup in \cite{RecUP}, we evaluate \sysname's effectiveness against attribute inference attacks \cite{RecUP} and membership inference attacks \cite{FLMIA} on LFW, comparing with Soteria and RecUP \cite{RecUP}.
As shown in Figure \ref{fig_aatr_mia}, \sysname achieves comparable model utility while significantly reducing the attack success rate (ASR) across both attack scenarios.}

\section{Evaluation in Malicious Threat Model}
\label{eval_loki}

\begin{figure}
    \centering
    \includegraphics[width=0.5\linewidth]{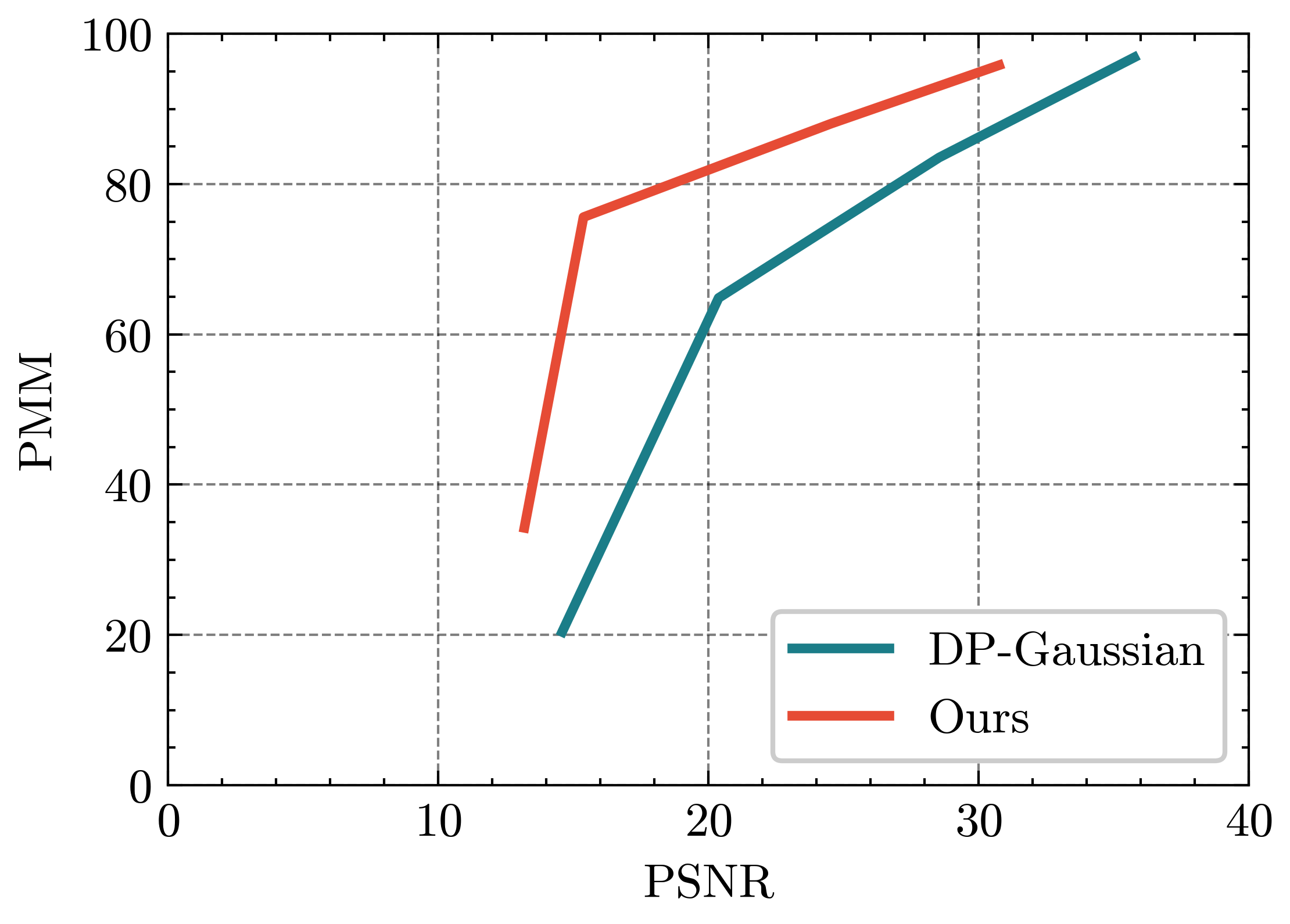}
    \caption{\revise{Defense effectiveness of \sysname against LOKI \cite{ZhaoSEEAB24} on CIFAR-10.}}
    \label{fig_loki}
\end{figure}

\revise{We evaluate \sysname's performance under a malicious threat model, where the server is allowed to modify model parameters, as considered in \cite{ZhaoSEEAB24}.
Following the configuration from \cite{ZhaoSEEAB24}, we conduct experiments on CIFAR-10 with a batch size of 16 and 4 local training steps per round.
As shown in Figure \ref{fig_loki}, \sysname demonstrates superior performance compared to DP against LOKI \cite{ZhaoSEEAB24}.
Specifically, while maintaining approximately 80\% of PMM, \sysname achieves a PSNR of around 14, whereas DP achieves a PSNR of 25, underscoring the effectiveness of \sysname.}

\section{Proof of Theorem \ref{convergence_thm}}
\label{proof_of_convergence}

We assume $|| \nabla_{\theta} \mathcal{L}(F_{\theta}(x),y) - \nabla_{\theta} \mathcal{L}(F_{\theta}(x^*),y)|| \leq \epsilon$.
By leveraging Assumption \ref{assumpt3}, there is:
\begin{equation}
\label{proof_convergence_our_1}
\begin{split}
&\mathbb{E} || \nabla_{\theta} \mathcal{L}(F_{\theta}(x_i),y_i) - g_{i}^{full} ||_2^2 \\
&= \mathbb{E} || \nabla_{\theta} \mathcal{L}(F_{\theta}(x_i),y_i) - \nabla_{\theta} \mathcal{L}(F_{\theta}(x^*),y) \\
&+ \nabla_{\theta} \mathcal{L}(F_{\theta}(x^*),y) - g_{i}^{full} ||_2^2 \\
&\leq \mathbb{E} || \nabla_{\theta} \mathcal{L}(F_{\theta}(x_i),y_i) - \nabla_{\theta} \mathcal{L}(F_{\theta}(x^*),y) ||_2^2 \\
&+ \mathbb{E} || \nabla_{\theta} \mathcal{L}(F_{\theta}(x^*),y) - g_{i}^{full} ||_2^2 = \sigma_i^2 + \epsilon^2.
\end{split}
\end{equation}

Based on Assumption \ref{assumpt4}, we obtain:
\begin{equation}
\label{proof_convergence_our_2}
\begin{split}
\nonumber
&\mathbb{E}||\nabla_{\theta} \mathcal{L}(F_{\theta}(x_i^*),y_i)||_2^2 \\
&= \mathbb{E}||\nabla_{\theta} \mathcal{L}(F_{\theta}(x_i^*),y_i) - \nabla_{\theta} \mathcal{L}(F_{\theta}(x_i),y_i) + \nabla_{\theta} \mathcal{L}(F_{\theta}(x_i),y_i)||_2^2 \\
&\leq \mathbb{E}||\nabla_{\theta} \mathcal{L}(F_{\theta}(x_i^*),y_i) - \nabla_{\theta} \mathcal{L}(F_{\theta}(x_i),y_i) ||_2^2 \\
&+ \mathbb{E}||\nabla_{\theta} \mathcal{L}(F_{\theta}(x_i),y_i)||_2^2
= G^2 + \epsilon^2.
\end{split}
\end{equation}

Applying the above two equations in Theorem 2 in \cite{convergence_non_iid} makes:
$\mathbb{E}[\mathcal{L}(F_{\theta}(\cdot),\cdot)] - \mathcal{L}(F_{\theta^*}(\cdot),\cdot) \leq \frac{2 \kappa}{\gamma + N}(\frac{Q+C}{u} + \frac{u \gamma}{2} \mathbb{E} ||\theta_{initial} - \theta^*||_2^2 ).$

\section{Projection Algorithm}
\label{projection_algorithm}

Given the gradient that clients upload, denoted as $g_{upload}$, our objective is to ensure that its distance from ground-truth gradients $g=\nabla_{\theta} \mathcal{L}(F_{\theta}(x),y)$ does not exceed $\epsilon$:
\begin{equation}
\begin{split}
&\min ||g_{upload} - g^*||_2^2, \\
|| g_{upload} - &g ||_2^2 \leq \epsilon^2, g^*=\nabla_{\theta} \mathcal{L}(F_{\theta}(x^*),y).
\end{split}
\end{equation}
For this optimization problem, there are two cases.
If $|| g_{upload} - g ||_2^2 \leq \epsilon^2$, the optimal solution obviously is $g_{upload} = g^*$.
Otherwise, it is $ g_{upload} = g + \epsilon \cdot \frac{g^*-g}{|| g^*-g ||_2}$.
In our experiment, clients check for $|| g_{upload} - g ||_2^2 \leq \epsilon^2$.
If it's satisfied, them upload $g^*$.
Otherwise, it makes slight modifications to $g^*$, \ie, uploading $ g_{upload} = g + \epsilon \cdot \frac{g^*-g}{|| g^*-g ||_2}$.

\begin{table}[!t]
\caption{The distance between $x$ and corresponding recovered data using four different attack methods.}
\label{privacy_supplementray_exp}
\centering
\scriptsize
\begin{tabular}{@{}ccccc@{}}
\toprule
Attack & iGLA & GradInvertion & InvertingGrad  \\ \midrule
$\sigma$ (Euclidean distance) & 0.00018 & 0.00053 & 0.00046  \\
$\sigma$ (Our evaluation network) & 0.003 & 0.007 & 0.005  \\
\bottomrule
\end{tabular}
\end{table}

\begin{table}[!t]
\caption{The similarity between $x$ and reconstructed data.}
\label{adaptive_attack_robustness}
\centering
\scriptsize
\begin{tabular}{@{}ccc@{}}
\toprule
              & PSNR    & Our evaluation network \\ \midrule
iDLG          & 10.6742 & 0.9847                 \\
InvertingGrad & 10.4378 & 0.9785                 \\ \bottomrule
\end{tabular}
\end{table}

\section{Evidence Supporting Section \ref{theory_privacy_part}}
\label{appendix_privacy_theoy_exp}

We here empirically demonstrate that $ \sigma \ll \text{dist}(x^*,x)$.
When employing Euclidean distance, $\sigma$ represents the MSE distance between the image $x$ and the data reconstructed from the gradients associated with $x$.
In the context of our evaluation network, $\sigma$ denotes the noise discrepancy between the original image $x$ and its gradient-based reconstruction.
We randomly extract 1000 from the training set of CIFAR10 as $x$.
We utilize LeNet and follow the same configuration as in Section \ref{evaluation_setup}.
Table \ref{privacy_supplementray_exp} reports the results averaged over 1000 samples using four different attack methods.
We further construct robust data for these 1000 samples.

The MSE distance and noise difference, averaged over these 1000 samples, between $x$ and their robust data are 0.0839 and 0.6258, significantly larger than $\sigma$, thus indicating the effectiveness of privacy protection analysis.

\section{Evaluation Network as Distance Metric}
\label{appendix_proof_evaluation_network_distance_metric}

Our evaluation network quantifies the amount of noise present in given data, and we define a distance metric as $ \text{dist}(x,y) = |D(x) - D(y)| $, where $ D(\cdot) $ denotes the predicted noise ratio.
This definition inherently ensures the non-negativity.
Furthermore, there is:
\begin{equation}
\label{evaluation_network_demonstration}
\begin{split}
\nonumber
&dist(x,z) = |D(x) - D(z)| = |D(x) - D(y) + D(y) - D(z)| \\
&\leq |D(x) - D(y)| + |D(y) - D(z)| = dist(x,y) + dist(y,z).
\end{split}
\end{equation}
Since our derivation only requires non-negativity and the triangle inequality (identity is not enforced), it qualifies as a valid metric for distance in this paper.

\section{Further Security Analysis}
\label{further_security_analysis}

In Section \ref{security_discussion}, we show that the equation $||g^* - g||_2^2 = const$ is underdetermined.
Here, we engage in an empirical analysis.
Specifically, we employ gradient optimization algorithm to solve $||g^* - g||_2^2 = const$ and denote the results as $g'$.
We then apply GLAs to reverse $g'$.
Notice that, in practice, we do not omit the weight factors.
Table \ref{adaptive_attack_robustness} reports the average PSNR between the reconstructed data from $g'$ and $x$ across 1000 samples.
As observed, the values of PSNR are typically low, indicating a significant difference between the reconstructed data and $x$.
Thus, \sysname is capable of effectively withstanding adaptive attacks.

\begin{figure}[!t]
    \centering
    \includegraphics[width=0.5\textwidth]{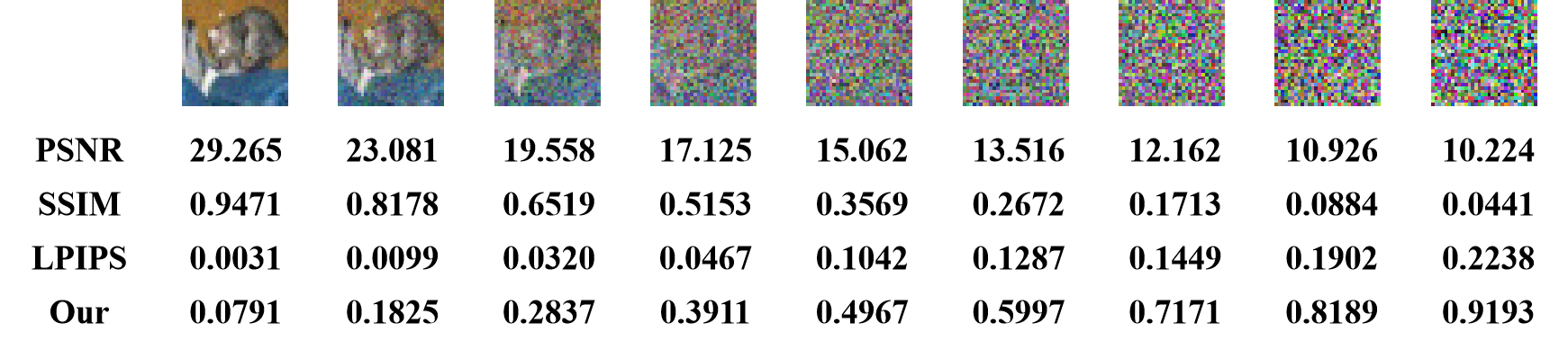}
    \caption{The relationship between different metric values and the degree of privacy leakage in the images.}
    \label{example_connection_metric_privacy}
\end{figure}

\section{Concrete Formula of Evaluation Metrics}
\label{concrete_formula_assess_privacy}

Figure \ref{example_connection_metric_privacy} illustrates how metric values correlate with privacy leakage.
For our evaluation network, a threshold of 0.4 is used to distinguish private vs. non-private outputs.
For PSNR, SSIM, and LPIPS, threshold values for safeguarding privacy are 15, 0.5, and 0.05, respectively.

\section{Overhead of \sysname in Edge Device Chips}
\label{appendix_overhead_edge_device}

We evaluate \sysname's computational overhead on four edge devices (Jetson Nano, Coral USB, Neural Stick, Coral PCIe) using ResNet50 with $224 \times 224$ input.
Based on \cite{edge_device}, a single forward pass takes 28.47 ms, 37.09 ms, 38.42 ms, and 44.04 ms, respectively.
Assuming backward passes match forward durations and the evaluation network is ResNet50-sized, solving Equation \ref{optim_task} with a batch size of 32 and 10 iterations (640$\times$ forward passes) yields total runtimes of 18.22s, 23.74s, 24.59s, and 28.19s.
While non-trivial, these costs remain practical for real-world deployment.

\end{document}